\documentclass{article}

\usepackage{graphicx}              
\usepackage{amsmath}               
\usepackage{amsfonts}              

\newcommand{\argmax}{\operatornamewithlimits{argmax}}

\begin{document}



\title{Online Algorithms For Parameter Mean And Variance Estimation In Dynamic Regression Models}

\author{Carlos A. G\'{o}mez-Uribe \\
cgomez@netlfix.com \\
Netflix, Inc. Los Gatos, CA USA}

\maketitle

\begin{abstract}
We study the problem of estimating the parameters of a regression model from a set of observations, each consisting of a response and a predictor. The response is assumed to be related to the predictor via a regression model of unknown parameters. Often, in such models the parameters to be estimated are assumed to be constant. Here we consider the more general scenario where the parameters are allowed to evolve over time, a more natural assumption for many applications. We model these dynamics via a linear update equation with additive noise that is often used in a wide range of engineering applications, particularly in the well-known and widely used Kalman filter (where the system state it seeks to estimate maps to the parameter values here). We derive an approximate algorithm to estimate both the mean and the variance of the parameter estimates in an online fashion for a generic regression model. This algorithm turns out to be equivalent to the extended Kalman filter. We specialize our algorithm to the multivariate exponential family distribution to obtain a generalization of the generalized linear model (GLM). Because the common regression models encountered in practice such as logistic, exponential and multinomial all have observations modeled through an exponential family distribution, our results are used to easily obtain algorithms for online mean and variance parameter estimation for all these regression models in the context of time-dependent parameters. Lastly, we propose to use these algorithms in the contextual multi-armed bandit scenario, where so far model parameters are assumed static and observations univariate and Gaussian or Bernoulli. Both of these restrictions can be relaxed using the algorithms described here, which we combine with Thompson sampling to show the resulting performance on a simulation. 
\end{abstract}

%
%

\section{Introduction}
Regression models are one of the main tools of statistical modeling and supervised machine learning. In regression models, \textit{responses} are related to \textit{predictors} by a probabilistic model that is a function of model parameters that we want to estimate from the data. The most common regression model, linear regression, assumes the response follows a Gaussian distribution that can take values anywhere in the real numbers. However, different applications have a response of a different nature. For example, the response might take values only in the positive real numbers (e.g., the time between the arrivals of two buses), or in the non-negative integers (e.g., the number of thunders in a day, or the number and identity of the items chosen by someone from a catalog of items). In such cases, assuming that the response is Gaussian might produce an inferior model to one obtained assuming a different distribution that better matches the nature of the response data. When the response takes on two values, it may be modeled as a Bernoulli random variable. When it is a non-negative real number, it may be modeled as an exponential, Erlang, or gamma distribution. When the response is a non-negative integer it may be modeled as a Poisson, negative binomial, or binomial random variable. When the response is a category it may be modeled by a multinomial or categorical distribution.  

All these distributions, and more, are part of the so-called exponential family (\cite{morris1982natural}, \cite{morris1983natural}, \cite{wainwright2008graphical}). The exponential family also includes distributions for random vectors with vector entries that are correlated, e.g., as in the multivariate Gaussian distribution, as well as with independent entries with different distribution types. The generalized linear model (GLM) introduced in \cite{nelder1972glm} provides the theory to build regression models with static parameters where the response follows a distribution in the exponential family. The GLM can then be seen as a generalization of linear regression. 

Many applications of regression models, indeed those that rely on the GLM, assume the parameters are static. This assumption is too restrictive when modeling time-series, which often exhibit trends, seasonal components, and local correlations. Similarly, in many applications the model parameters represent variables that describe a portion of the state of the underlying system that cannot be or are not directly measured, but with well-known relationships that describe their dynamics. The response can then be thought of as a noisy function of the system state and the predictors, and the goal of the regression model is to estimate the system state over time based on the observation time-series.

Situations where an underlying dynamical system is only observed through noisy measurements are often encountered in engineering applications, and the celebrated and widely used Kalman filter, introduced in \cite{kalman1960}  solves the corresponding regression model when the parameter dynamics (linear with additive Gaussian noise) and the observation distribution (Gaussian with a mean that is a linear function of the state) are simple enough. The Kalman filter can be seen as an algorithm that efficiently estimates the mean and covariance matrix of the parameters in a linear regression model, where the parameters evolve in time through linear dynamics, based on the time-series of responses and predictors. Furthermore, the Kalman filter is an online algorithm that updates the parameter estimates in a relatively simple fashion when new observations arrive, without having to re-analyze all previous data. In addition, it is a second-order algorithm, in contrast to stochastic gradient descent (e.g., \cite{bottou2010large}). So while the Kalman filter takes more computation per observation than stochastic gradient descent, it can converge to an accurate estimate of the parameters with fewer observations, while allowing for much more flexibility in the modeling of the underlying parameter dynamics. Lastly but importantly, because the Kalman filter provides an estimate of the mean and covariance matrix, unlike other approaches that only focus on the mean, its results can be used to construct confidence intervals or samples of the parameters or of the model predictions --- these statistics are often necessary in several applications such as in contextual multi-armed bandits (e.g., \cite{li2010contextual}). Many generalizations to the Kalman filter have now been developed, e.g., to allow for non-linear state dynamics state dynamics, or for observations that are noisy and non-linear and/or non-Gaussian functions of the state as in the extended Kalman filter, e.g., see \cite{simon2006optimal} for a good overview.

More recently, there has been interest in merging the ideas from Kalman filtering, namely modeling dynamics in the parameters of a regression model, with the flexibility to model observations from the wide class of probability distributions that the GLM allows through its treatment of the exponential family (\cite{west1985dynamic},  \cite{fahrmeir1992posterior}, \cite{durbin2000time}, \cite{durbin2012time}, \cite{klein2003state}, \cite{harrison1999bayesian}). It turns out that approximate algorithms that are very similar to the extended Kalman filter can be derived for a wide range of choices for the observation distributions. Introducing and describing some of these algorithms for a fairly general class of models, including the multivariate GLM, is the main focus of this paper. The derivation we follow is novel and simpler (e.g., it does not invoke any Kalman filter theory nor uses conjugacy). And the form of the exponential family we use is slightly more general than that used in other references on these methods, because the nuissance parameter matrix $\mathbf{\Phi}$ included in our model is absent in other references that deal with multivariate responses. 

The second focus of this paper is to propose the application of these methods to the contextual multi-armed bandit problem (\cite{li2010contextual}, \cite{chapelle2011empirical}), and show the resulting performance through simulations. This application is novel to the best of our knowledge. It broadens the class of models that can be used to describe the rewards, separates the concept of a response and a reward, and allows for the explicit modeling of dynamics in the parameters that map the context to the rewards. 

Section \ref{sec:setup} introduces the class of models we study, and reviews the exponential family and the generalized linear model. Section \ref{sec:algo} describes the general online algorithm to estimate the mean and variance of the parameters, and specializes it to the multivariate exponential family, the univariate exponential family, and to several examples of commonly used distributions. We sketch the derivation of the algorithm in Section \ref{sec:derivation}. We apply the methods developed to the contextual multi-armed bandit problem in Sections \ref{contextualBandits}, and conclude in Section \ref{conclusion}.

\section{Model Setup} \label{sec:setup}

We assume all vector and matrix entries are real numbers. We denote vectors using boldface small caps, and assume them to be column vectors. We use small caps for scalars, and boldface large caps for matrices.  If $\mathbf{A}$ is a matrix, we denote its inverse by $\mathbf{A}^{-1}$, and its transpose by $\mathbf{A'}$.   

\subsection{Regression Models For The Response}
We assume we have received $n$ pairs of observations $(\mathbf{X_i}, \mathbf{\mathbf{y_i}})$, for $i=1, \ldots, n,$ where $\mathbf{y_i} \in \mathbb{R}^d$ is the $i$-th \textit{response} that we want to explain based on the $i$-th \textit{predictor}  $\mathbf{X_i} \in \mathbb{R}^k \times  \mathbb{R}^c$, for some positive integers $d$, $k$ and $c$. We denote by $D_n$ the information or history after $n$ observations, i.e., $D_n$ is the set that includes the first $n$ responses and predictors.

We postulate a model that relates the response to the predictor via a probability distribution $p(\mathbf{\mathbf{y_i}}|\mathbf{X_i}, \mathbf{\theta_i}, \mathbf{\Phi_i} ),$ where $\mathbf{\theta_i} \in \mathbb{R}^k$ are the model parameters at the  $i$-th observation, with one parameter for each row in the predictor matrix $\mathbf{X_i}.$ $\mathbf{\Phi_i}$ is a $d$-by-$d$ matrix and nuisance parameter that is assumed known and that we will often omit. As we will see later, the nuisance parameter plays the role of the covariance of the observations in linear regression, and is the identity matrix in many other cases of interest.

We consider regression models where the probability of the response depends on the predictors and the parameters only through the $c$-by-$1$ \textit{signal} $\mathbf{\lambda_i}=\mathbf{X'_i}\mathbf{\theta_i},$ namely, models where $p(\mathbf{\mathbf{y_i}}|\mathbf{X_i}, \mathbf{\theta_i})=p(\mathbf{y_i}|\mathbf{\lambda_i}).$ 

Rather than working with $p(\mathbf{y_i}|\mathbf{\lambda_i})$, we will typically work with its logarithm, denoted by $l(\mathbf{y_i}|\mathbf{\lambda_i}),$ which we assume to be a well-behaved function, particularly that the first and second derivatives with respect to $\mathbf{\lambda_i}$ exist and are finite.

Many of the commonly used regression models fit the model above, including univariate and multivariate linear regression with response $\mathbf{y} \sim  \mathcal{N}(\mathbf{\lambda},\mathbf{\Sigma}),$ logistic or binomial, categorical or multinomial, exponential, Poisson, negative binomial, gamma, etc. Our model also includes cases where  the different entries in the response vector are conditionally independent given the signal, and follow any distribution that is a function of the signal or a subset of the signal entries. For example, we can have as many predictor vectors as entries in $\mathbf{y}$, i.e., $c=d,$ and have the $j$-the response entry depend only on the $j$-th signal entry so that $p(\mathbf{y_i}|\mathbf{\lambda_i})=\prod_jp_j(\mathbf{y_i}_j|\mathbf{\lambda_i}_j)$, with $p_j()$ being a different function for different response entries $j$. Because all the entries still depend on the parameters $\mathbf{\theta},$ this setup allows for combining the information from different types of measurements that depend on the parameters to obtain more accurate parameter estimates. Also, the predictor matrix $\mathbf{X_i}$ may have a lot of structure, e.g., to allow for some parameters to be shared across entries in the response vector, and others that are specific to subsets of the response.

\subsection{The Natural Exponential Family} \label{sec:regModelsNEF}
Here we drop the time subscript of our vectors and matrices to avoid notational clutter, so, e.g., $\mathbf{y_i}$ becomes simply $\mathbf{y}$. All of the probability distributions of interest to model the response $\mathbf{y}$ mentioned above, and more, can be re-arranged so that their log-likelihood has the following so-called \textit{natural exponential form}
\begin{align}
l \big( \mathbf{y} | \mathbf{\eta}, \mathbf{\Phi}\big) = & \mathbf{\eta' \Phi^{-1} y} - b\big(\mathbf{\eta, \Phi} \big)+c\big(\mathbf{y, \Phi} \big). \label{eq:nef}
\end{align}
Here, $\mathbf{\eta}$  is a $d$-by-1 vector referred to as the natural parameter, with $\eta_j$ in its $j$-th entry. It is a function of the signal $\mathbf{\lambda}$ in our models --- this will be made specific in the next section. Crucially, the function $b()$ is independent of $\mathbf{y}$ and the function $c()$ is independent of $\mathbf{\eta}.$ We assume that $b()$ is twice differentiable with respect to its argument. The $d$-by-$d$ nuisance parameter matrix $\mathbf{\Phi}$ is assumed symmetric and known. When $\mathbf{\Phi}$ is unknown, it can be estimated through several methods that are not online,  e.g., see Chapter 13 in \cite{durbin2012time}.

It can be shown, e.g., see Appendix \ref{sec:meanVarNEF}, that the mean and covariance matrix of $\mathbf{y}$ are given by 
\begin{align}
\mathbf{\mu(\eta)} =&E[\mathbf{y}] =  \mathbf{\Phi} \frac{\partial b}{\partial \mathbf{\eta}}   \label{eq:nefMean} \\ 
 \mathbf{\Sigma_y(\eta)}=&E[(\mathbf{y}- \mathbf{\mu(\eta)})(\mathbf{y}- \mathbf{\mu(\eta)})'] =  \mathbf{\Phi} \frac{\partial^2 b}{\partial \mathbf{\eta}^2} \mathbf{\Phi}, \label{eq:nefVar}
\end{align}
where $\frac{\partial b}{\partial \mathbf{\eta}}$ is a column vector with $ \frac{\partial b}{\partial \mathbf{\eta_j}}$ in its $j$-th entry, and $\frac{\partial^2 b}{\partial \mathbf{\eta}^2}$ is the $d$-by-$d$ matrix with $\frac{\partial^2 b}{\partial \eta_i \partial \eta_j}$ in its $i$-th row and $j$-th column.

Most of the frequently encountered univariate and multivariate distributions have the exponential form above. In addition, a union of independent random vectors that are in the natural exponential family is also in the natural exponential family. E.g., a random vector with entries distributed according to different members of the exponential family is still in the natural exponential family with a natural parameter given by the union of the natural parameters of its entries. This will allow us to estimate shared parameters in a regression model from multiple time series of a potentially different nature.

We consider models where the response is distributed according to Equation \ref{eq:nef}, which is a function of $\mathbf{\eta}$. We use the GLM to relate $\mathbf{\eta}$ to the signal $\mathbf{\lambda}.$ 

\subsection{The Generalized Linear Model}\label{sec:glm}

In the GLM, introduced in \cite{nelder1972glm}, we assume that the signal $\mathbf{\lambda}$ is a function of the mean $\mathbf{\mu}$ of the observation  $\mathbf{y}$, in addition to assuming Equation \ref{eq:nef} for the observation $\mathbf{y}$. The GLM then assumes that there is a known one-to-one mapping between the natural parameter $\mathbf{\eta}$  in Equation \ref{eq:nef} and the signal $\mathbf{\lambda},$ so we can view the likelihood or any statistic of  $\mathbf{y}$ either as a function of $\mathbf{\eta}$ or of $\mathbf{\lambda}$. Specifically, we have
\begin{equation}
\mathbf{\lambda}=\mathbf{X'\theta}=\mathbf{g(\mu)}=\mathbf{u(\mathbf{\eta})}=\mathbf{v^{-1}}(\mathbf{\eta}), \label{eq:glm1}
\end{equation}
for known functions $\mathbf{u()},$ $\mathbf{v()}$ and $\mathbf{g()}$, and where the so-called \textit{link function} $\mathbf{g()}$ maps the mean $\mathbf{\mu}$ of $\mathbf{y}$ to the signal. The mean $\mathbf{\mu}$ and covariance $\mathbf{\Sigma_y}$ in Equations \ref{eq:nefMean} and \ref{eq:nefVar} can then be seen to be either a function of the natural parameter $\mathbf{\eta}$ or of the signal $\mathbf{\lambda,}$ e.g.,
\begin{align}
\mathbf{\mu}= & \mathbf{\Phi} \frac{\partial b}{\partial \mathbf{\eta}} = \mathbf{\tau}(\mathbf{\eta}) =  \mathbf{\tau}(\mathbf{v(\mathbf{\lambda})} ) =  \mathbf{h}(\mathbf{\lambda}) . \label{eq:glm2}
\end{align}
Here the function $ \mathbf{h}()$ is the inverse of $\mathbf{g()}$. We refer to $\mathbf{h}(\mathbf{\lambda})$ as the \textit{response function}; it maps the signal to the mean of the response and plays a prominent role in our algorithms.

Any invertible function $\mathbf{g()}$ can be used as the link function in a GLM, but one that often makes sense, and where the mathematics to learn the model simplifies, is the \textit{canonical link} that results in the signal being equal to the natural parameter, i.e., in $\mathbf{\lambda}=\mathbf{\eta}=\mathbf{u}(\mathbf{\eta}).$ 

Previous treatments of the GLM in the context of dynamic parameters consider either a univariate response with the univariate case of Equation \ref{eq:nef} (e.g., see \cite{west1985dynamic}), or a multivariate response with Equation \ref{eq:nef} but with the nuissance parameter matrix $\mathbf{\Phi}$ equal to the identity (e.g., see Chapter 2 in \cite{klein2003state}). In this sense our treatment is a slight generalization.

\subsection{Parameter Dynamics: The Kalman Filter}
We assume that the parameters evolve according to
\begin{equation}
\mathbf{\theta_{t}}=\mathbf{G_t\theta_{t-1}} + \mathbf{B_t u_{t-1}}+\mathbf{\omega_t}, \label{eq:parDynamics}
\end{equation}
where $\mathbf{\omega_t} \in \mathbb{R}^k$ is a zero-mean random vector with known covariance matrix $\mathbf{W_t}.$ We also assume that the \textit{noise} $\mathbf{\omega_t}$ is uncorrelated with itself over time, and uncorrelated with the observation parameters. The known \text{input} vector $\mathbf{u_t}$ drives the parameters through the appropriately sized and known matrix $\mathbf{B_t},$ and $\mathbf{G_t}$ is a known $k$-by-$k$ square matrix. We also assume that at time zero the mean and variance of $\mathbf{\theta}$ are known, i.e., that $\mathbf{\theta_{0}} \sim (\mathbf{m_0}, \mathbf{C_0}).$ The general setup above includes several special cases of interest, described next.

When $\mathbf{G_t}=\mathbf{I}$ (the identity matrix), $\mathbf{B_t u_t}=\mathbf{0}$, and $\mathbf{C_0}=\mathbf{0},$ we end up with the simple parameter dynamics $\mathbf{\theta_{t+1}}=\mathbf{\theta_t},$ which is the standard regression problem with static parameters. This becomes the GLM when we also assume that the response is in the exponential family, with natural parameter that is a function of the signal. In this sense, our model is a generalization of the GLM  where the parameters are allowed to vary over time.

When $\mathbf{G_t}=\mathbf{I}$ and $\mathbf{B_t u_t}=\mathbf{0}$, we end up with the simple parameter dynamics $\mathbf{\theta_{t+1}}=\mathbf{\theta_t}+\mathbf{\omega_t}$. This allows the parameters to drift or diffuse over time in an unbiased (zero-mean) way, according to the noise $\mathbf{\omega_t}$. This model is appealing for a range of applications, e.g., it could model the conversion rate of visitors to the Netflix signup page as a function of their browser, country, day of week, time of day, etc. 

Equation \ref{eq:parDynamics} is central to the study of linear dynamical systems, control theory, and other related areas. Specifically, it is core to the Kalman filter, which also assumes a dynamical system that evolves according to Equation \ref{eq:parDynamics}, but with a response that is a linear function of the parameters (the state in Kalman filter parlance) and additive Gaussian noise. The Kalman filter is an online algorithm that estimates the mean and variance of the system state from the noisy observations. So our setup is very related. The main difference is that we do not restrict our response to be Gaussian, but rather a non-linear function of  $\mathbf{\lambda}$, itself a linear function of the state. The non-linearity and the noise characteristics of the observations follow from the choice of regression model made, e.g., from the specific mapping between the signal and the response: choosing a Gaussian distribution for the response with a mean equal to the signal yields the standard Kalman filter. In this sense, our model is a generalization of the standard Kalman filter. A variant of the Kalman filter known as the extended Kalman filter deals with general non-linear observations, and has been applied to responses modeled through the exponential family (\cite{durbin2012time}, \cite{fahrmeir1992posterior}), yielding an algorithm that can be shown to be equivalent to ours.

Based on the assumptions above, we obtain the following factorization of the joint probability function of predictors, responses and parameters.
\begin{equation}
 p(\mathbf{X_1, \ldots, X_t}) p(\mathbf{\theta_0}) \biggr(\prod_{i=1}^t p(\mathbf{\theta_{i}}|\mathbf{\theta_{i-1}}) p(\mathbf{y_i}|\mathbf{\lambda_i})\biggr).
\label{eq:jointMaster}
\end{equation}
\section{Estimating Model Parameters} \label{sec:algo}
We seek an algorithm to compute the mean and covariance matrix of the model parameters using all the observations we have at any given time. We want this algorithm to be online, i.e., to perform a relatively simple update to the previous mean and covariance estimates when a new observation arrives, without having to re-analyze previous observations.
\subsection{The General Algorithm}
Let $\mathbf{m_t}$ and $\mathbf{C_t}$ denote the mean and covariance matrix of $\big(\mathbf{\theta_{t}}|D_t\big).$ First, we initialize the mean and covariance of $\mathbf{\theta_0}$ to the known values $\mathbf{m_0}$ and $\mathbf{C_0}$. 

We proceed by induction: we assume we know that $\big(\mathbf{\theta_{t-1}}|D_{t-1}\big)$ has parameter mean and covariance matrix $\mathbf{m_{t-1}}$ and $\mathbf{C_{t-1}}$,  and use them and the new observation to compute $\mathbf{m_{t}}$ and $\mathbf{C_{t}}$ through a two stepped process suggested by the following simple relation:
\begin{align}
p\big(\mathbf{\theta_{t}}|D_t\big) & =  \ p\big(\mathbf{\theta_{t}}|D_{t-1}, \mathbf{X_t}, \mathbf{y_t} \big) \propto   p\big(\mathbf{\theta_{t}}, \mathbf{X_t}|D_{t-1}\big) p\big(\mathbf{y_t}|\mathbf{X_t},  \mathbf{\theta_{t}} \big)  \nonumber \\
 & =  \ p\big(\mathbf{\theta_{t}} |D_{t-1}\big) p\big(\mathbf{X_{t}} |D_{t-1}\big) p\big(\mathbf{y_t}|\mathbf{X_t},  \mathbf{\theta_{t}} \big) \nonumber \\
& \propto   \ p\big(\mathbf{\theta_{t}} |D_{t-1}\big) p\big(\mathbf{y_t}|\mathbf{\lambda_t} \big), \text{ so that}  \nonumber 
\end{align}
\begin{align}
\log{\big(p\big(\mathbf{\theta_{t}}|D_t\big)\big)} &=  \  \log{\big(p\big(\mathbf{\theta_{t}}|D_{t-1}\big)\big)} +  l(\mathbf{y_t}|\mathbf{\lambda_t}) + \text{const}. \label{eq:bayes} 
\end{align}
Equation \ref{eq:bayes} relates $\big(\mathbf{\theta_{t}}|D_t\big)$ to its \text{prior} $\big(\mathbf{\theta_{t}}|D_{t-1}\big),$ which predicts $\mathbf{\theta_{t}}$ based on all previous information up to but ignoring the observation at time $t$, and the log-likelihood of the latest observation $l(\mathbf{y_t}|\mathbf{\lambda_t})$. 

\subsubsection{Step 1: Prediction}
We compute the mean and covariance of the prior $\big(\mathbf{\theta_{t}}|D_{t-1}\big)$ via
\begin{align}
\mathbf{a_t} = &  \mathbf{G_t m_{t-1}} + \mathbf{B_t u_{t-1}}, \text{ and}  \label{eq:meanPred} \\
\mathbf{R_t} = & \mathbf{G_tC_{t-1}G_t'} + \mathbf{W_t}. \label{eq:varPred}
\end{align}
This equation is exact and does not require assuming any functional form for $\big(\mathbf{\theta_{t-1}}|D_{t-1}\big).$ It follows fairly directly from Equation \ref{eq:parDynamics}, e.g., see Appendix \ref{sec:meanVarPredictionApp} for a derivation of these and other equations in this section. When the parameter dynamics are non-linear, the mean and covariance of $\big(\mathbf{\theta_{t}}|D_{t-1}\big)$ can be approximated through expressions identical to Equations \ref{eq:meanPred} and \ref{eq:varPred} by suitably re-defining the matrices that appear in them, e.g., by linearizing the parameter dynamics around $\mathbf{m_{t-1}},$ or via numerical simulation as in the so-called unscented Kalman filter (\cite{durbin2012time}, \cite{klein2003state} and \cite{simon2006optimal}).

Note that when $\mathbf{G_t}$ is the identity matrix, Equation \ref{eq:varPred} shows that the variance of the parameter estimates always increases in the prediction step, unless $\mathbf{W_t}=\mathbf{0}.$ In addition, because the system input $\mathbf{u_t}$ is deterministic, it does not contribute to the covariance matrix.

When the predictor $\mathbf{X_t}$ becomes known, we can use Equations \ref{eq:meanPred} and \ref{eq:varPred} to determine the mean $\mathbf{f_t}$ and covariance matrix $\mathbf{\Omega_t}$ of the signal $\mathbf{\lambda_t}$ given $D_{t-1}$ and $\mathbf{X_t}$:
\begin{align}
\mathbf{f_t} = &  \mathbf{X_t' a_t}, \text{ and}  \label{eq:meanSignalPred} \\
\mathbf{\Omega_t} = & \mathbf{X_t' R_t X_t}. \label{eq:varSignalPred}
\end{align}

Lastly, the covariance matrix between the signal and the parameters is given by $\mathbf{X_t'R_t}$. The latter follows from the fact that signal is a linear function of the parameters, with $\mathbf{X_t'}$ as the weights.
\subsubsection{Step 2: Estimation}
Now we update the estimates from the prediction step to incorporate the new observation, obtaining the mean $\mathbf{m_t}$ and covariance $\mathbf{C_t}$ of the posterior $\big(\mathbf{\theta_{t}}|D_t\big).$ We first compute the following matrices
\begin{align}
\mathbf{Q_t} =& \biggr[-\frac{\partial^2  l(\mathbf{y_t}|\mathbf{\lambda_t=f_t})}{\partial \mathbf{\lambda_t}^2}\biggr]^{-1} + \mathbf{\Omega_t}, \text{ and } \label{eq:QMatrix} 
\mathbf{A_t} =& \mathbf{R_tX_tQ_t^{-1}}.
\end{align}
Here $\frac{\partial^2  l(\mathbf{y_t}|\mathbf{\lambda_t=f_t})}{\partial \mathbf{\lambda_t}^2}$ is the $c$-by-$c$ Hessian matrix of the log-likelihood, evaluated at the predicted value of the signal $\mathbf{f_t}.$ As we will see, in many models of interest, this matrix is the negative of the variance of $\mathbf{y_t}$ evaluated at the predicted signal value $\mathbf{f_t}.$ The matrix $\mathbf{Q_t}$ then grows with the expected variance of the predicted signal response $\mathbf{\Omega_t}$, but decreases when the expected variance of the response increases.

 We then compute the covariance $\mathbf{C_t}$ via:
\begin{align}
\mathbf{C_t} = & \mathbf{R_t} - \mathbf{A_tQ_tA_t'}  
= \mathbf{R_t} - \mathbf{R_tX_tQ_t^{-1}X_t'R_t}. \label{eq:varUpdate2} 
\end{align}
Computing the inverse of $\mathbf{Q_t}$ starting from Equation \ref{eq:QMatrix} can often be numerically unstable, e.g., because the determinant of $\frac{\partial^2  l(\mathbf{y_t}|\mathbf{\lambda_t=f_t})}{\partial \mathbf{\lambda_t}^2}$ can be very small in magnitude. A more robust way to compute $\mathbf{Q_t^{-1}}$ is via
\begin{align} -\frac{\partial^2  l(\mathbf{y_t}|\mathbf{f_t})}{\partial \mathbf{\lambda_t}^2}\biggr[ \mathbf{I}+ \mathbf{\Omega_t}\biggr(\mathbf{I} -\frac{\partial^2  l(\mathbf{y_t}|\mathbf{f_t})}{\partial \mathbf{\lambda_t}^2} \mathbf{\Omega_t}  \biggr)^{-1}\frac{\partial^2  l(\mathbf{y_t}|\mathbf{f_t})}{\partial \mathbf{\lambda_t}^2}   \biggr].\end{align}
This expression follows directly from Equations \ref{eq:varUpdate2} and \ref{eq:QMatrix} after applying the Kailath variant of the Woodbury identity (e.g., see \cite{petersen2008matrix}).

We finally compute the mean of the parameters by:

\begin{align}
\mathbf{m_t} =&  \mathbf{a_t}+ \mathbf{C_tX_t}  \frac{\partial  l(\mathbf{y_t}|\mathbf{\lambda_t=f_t})}{\partial \mathbf{\lambda_t}} \label{eq:meanUpdate} \\
=& \mathbf{a_t}+ \mathbf{A_t}\biggr(-\frac{\partial^2  l(\mathbf{y_t}|\mathbf{\lambda_t=f_t})}{\partial \mathbf{\lambda_t}^2}\biggr)^{-1} \frac{\partial  l(\mathbf{y_t}|\mathbf{\lambda_t=f_t})}{\partial \mathbf{\lambda_t}}. \nonumber
\end{align}

Our main algorithm proceeds by executing the prediction and estimation steps for each arriving observation, namely evaluating Equations \ref{eq:meanPred}, \ref{eq:varPred},  \ref{eq:varUpdate2} and \ref{eq:meanUpdate} with every new observation. Equations \ref{eq:meanUpdate} and \ref{eq:varUpdate2} are approximate, and follow from (1) a second-order Taylor expansion of $l(\mathbf{y_t}|\mathbf{\lambda_t})$ around $\mathbf{a_t}$, and (2) assuming the prior $\big(\mathbf{\theta_{t}}|D_{t-1}\big)$ is Gaussian with mean and covariance given by $\mathbf{a_t}$ and $\mathbf{R_t}$. A sketch of the argument is described in Section \ref{sec:derivation}. Because the two assumptions we make are exact in the case of linear regression with a Gaussian prior for $\mathbf{\theta_{0}},$ Equations \ref{eq:meanUpdate} and \ref{eq:varUpdate2} are exact in that case and correspond to the standard Kalman filter equations.  

West \textit{et al.} (\cite{west1985dynamic}) make the different approximation that the prior $\big(\mathbf{\theta_{t-1}}|D_{t-1}\big)$ is conjugate to the likelihood $ l(\mathbf{y_t}|\mathbf{\lambda_t}),$ obtaining a slightly different algorithm that has only been developed for the univariate response scenario.

\subsection{The Univariate Signal Case}
Many regression models involve a scalar signal $\lambda_t = \mathbf{x_t'}\mathbf{\theta_{t}}$ and a scalar response $y_t,$ where the predictor is now simply a vector $\mathbf{x_t}$. This is the situation for the most commonly encountered regression models, such as linear, logistic, Poisson or exponential. The matrices $\mathbf{\Omega_t},$ $\frac{\partial^2  l(y_t|\mathbf{\lambda_t})}{\partial \mathbf{\lambda_t}^2},$ and $\mathbf{Q_t}$ then also become scalars, so the update Equations \ref{eq:meanUpdate} and \ref{eq:varUpdate2} simplify to
\begin{align}
\mathbf{C_t} = & \mathbf{R_t} + \frac{\frac{\partial^2  l(y_t|\lambda_t=f_t) }{\partial \lambda_t^2}}{1 - \frac{\partial^2  l(y_t|\lambda_t=f_t) }{\partial \lambda_t^2}\mathbf{x_t'R_tx_t}} \mathbf{\big(R_tx_t\big)\big(x_t'R_t\big)}, \text{ and } \label{eq:varUpdateScalarSignal} \\ 
\mathbf{m_t} =&  \mathbf{a_t}+ \mathbf{C_tx_t}  \frac{\partial  l(y_t|\lambda_t=f_t)}{\partial \lambda_t}, \label{eq:meanUpdateScalarSignal}
\end{align}
where the predicted signal is $f_t=\mathbf{x_t'a_t}$. The result is very appealing because no matrix inverses need to be computed.

\subsection{The Dynamic Generalized Linear Model} \label{sec:dynGLM}
Here we consider models where the response is in the exponential family of Equation \ref{eq:nef}, and where the natural parameter $\mathbf{\eta}$ is related to the signal via Equations \ref{eq:glm1} and \ref{eq:glm2}. The gradient in these models can be shown to be given by
\begin{align}
\frac{\partial  l(\mathbf{y_t}|\mathbf{\lambda_t})}{\partial \mathbf{\lambda_t}} = & \frac{d \mathbf{\eta_t}'}{d \mathbf{\lambda_t}} \frac{\partial  l(\mathbf{y_t}|\mathbf{\eta_t})}{\partial \mathbf{\eta_t}} = \frac{\partial  \mathbf{h(\lambda_t})'}{\partial \mathbf{\lambda_t}} \mathbf{\Sigma_{y_t}^{-1}}(\mathbf{\lambda_t}) \big(\mathbf{y_t}-\mathbf{h(\lambda_t)}\big). \label{eq:gradientGLM} 
\end{align}
So the gradient is always proportional to the error $\mathbf{y_t}-\mathbf{h(\lambda_t)}$, the difference between the response and its mean according to the model at the given signal. The covariance matrix $\mathbf{\Sigma_{y_t}^{-1}}(\mathbf{\lambda_t})$ is in general a function of the signal $\mathbf{\lambda_t},$ but we drop that dependence in our notation below to reduce clutter.


The Hessian $ \frac{\partial^2  l(\mathbf{y_t}|\mathbf{\lambda_t})}{\partial \mathbf{\lambda_t}^2}$ is then obtained by differentiating Equation \ref{eq:gradientGLM} with respect to the signal once more, resulting in
\begin{align}
   \frac{\partial}{\partial  \mathbf{\lambda_t}}\biggr[\frac{\partial  l(\mathbf{y_t}|\mathbf{\lambda_t})}{\partial \mathbf{\lambda_t}}\biggr] =  \frac{\partial}{\partial \mathbf{\lambda_t}}\biggr[ \frac{\partial  \mathbf{h(\lambda_t})'}{\partial \mathbf{\lambda_t}}   \mathbf{\Sigma_{y_t}^{-1}} \biggr]  \big(\mathbf{y_t}-\mathbf{h(\lambda_t)}\big) \nonumber \\
  - \frac{\partial  \mathbf{h(\lambda_t})'}{\partial \mathbf{\lambda_t}}   \mathbf{\Sigma_{y_t}^{-1}} \frac{\partial  \mathbf{h(\lambda_t})}{\partial \mathbf{\lambda_t}}. \label{eq:HessianGLM}
 \end{align}
 Evaluating Equations \ref{eq:gradientGLM} and  \ref{eq:HessianGLM} for a given choice of the likelihood and link function (which determines the response function $\mathbf{h(\lambda_t)}$) at the predicted signal $\mathbf{\lambda_t = f_t}$, and plugging the resulting expressions into Equations \ref{eq:varUpdate2} and \ref{eq:meanUpdate} completes the algorithm. 
 
 \subsubsection{The canonical Link}
 When the canonical link is used, the natural parameter is equal to the signal, so $\frac{d \mathbf{\eta_t}}{d \mathbf{\lambda_t}}=\mathbf{I}$, and $\frac{\partial  \mathbf{h(\lambda_t})}{\partial \mathbf{\lambda_t}} = \mathbf{\Sigma_{y_t}\Phi_t^{-1}}$. Equations \ref{eq:gradientGLM} and \ref{eq:HessianGLM} simplify to
 \begin{align}
\frac{\partial  l(\mathbf{y_t}|\mathbf{\lambda_t})}{\partial \mathbf{\lambda_t}} = &  \mathbf{\Phi_t^{-1}} \big(\mathbf{y_t}-\mathbf{h(\lambda_t)}\big), \text{and} \\
 \frac{\partial^2  l(\mathbf{y_t}|\mathbf{\lambda_t})}{\partial \mathbf{\lambda_t}^2} =& - \mathbf{\Phi_t^{-1}}   \mathbf{\Sigma_{y_t}}(\mathbf{\lambda_t}) \mathbf{\Phi_t^{-1}}.
\end{align}
So the gradient is proportional to the error, as before, and the Hessian is proportional to the negative of the covariance of $ \mathbf{y_t}.$  Evaluating the gradient and Hessian above at the predicted signal $\mathbf{f_t}$, and plugging in the resulting expressions into Equations \ref{eq:varUpdate2} and \ref{eq:meanUpdate} yields the update equations
\begin{align}
\mathbf{Q_t} =& \mathbf{\Phi_t}   \mathbf{\Sigma_{y_t}^{-1}(f_t)} \mathbf{\Phi_t} + \mathbf{X_t'R_tX_t},  \\
\mathbf{A_t} =& \mathbf{R_tX_tQ_t^{-1}}, \\
\mathbf{C_t} = & \mathbf{R_t} - \mathbf{A_tQ_tA_t'}= \mathbf{R_t} - \mathbf{R_tX_tQ_t^{-1} X_t'R_t}  \\
=& \mathbf{R_t} - \mathbf{R_tX_t}\biggr[ \mathbf{E_t} -  \mathbf{E_t\Omega_t}\biggr( \mathbf{I} + \mathbf{E_t\Omega_t} \biggr)^{-1}  \mathbf{E_t}\biggr] \mathbf{X_t'R_t}  \label{eq:meanUpdatecanonicalNEFVar2} \\
\mathbf{m_t} =&  \mathbf{a_t}+ \mathbf{C_tX_t} \mathbf{\Phi_t^{-1}} \big(\mathbf{y_t}-\mathbf{h(f_t)}\big),   \text{ where} \label{eq:meanUpdatecanonicalNEF}\\
 \mathbf{E_t}=& \mathbf{\Phi_t^{-1} \Sigma_{y_t}(f_t) \Phi_t^{-1}}.
\end{align}
Equation \ref{eq:meanUpdatecanonicalNEFVar2} is the numerically stable analog of Equation \ref{eq:varUpdate2} that avoids inverses of potentially close-to-singular matrices.

\textbf{Multivariate linear regression} is one of many examples that falls in this class of models. There, $\mathbf{y_t} \sim \mathcal{N}(\mathbf{\lambda_t, \Sigma}),$ and $\mathbf{y_t}$ can be shown to be in the natural exponential family (e.g., see Equation \ref{eq:GaussianNEF} in the Appendix), with $\Phi_t = \mathbf{\Sigma}$ and $\mathbf{h(\lambda_t)=\lambda_t},$ and covariance matrix equal to $\mathbf{\Sigma},$ which in this case is independent of the signal. So the equations above yield the standard Kalman filter equations.
\begin{align}
\mathbf{Q_t} =& \mathbf{\Sigma} + \mathbf{X_t'R_tX_t},  \\
\mathbf{A_t} =& \mathbf{R_tX_tQ_t^{-1}}, \\
\mathbf{C_t} = & \mathbf{R_t} - \mathbf{A_tQ_tA_t'} \text{, and } \nonumber \\
\mathbf{m_t} =&  \mathbf{a_t}+ \mathbf{C_tX_t} \mathbf{\Sigma^{-1}} \big(\mathbf{y_t}-\mathbf{f_t}\big).   \label{eq:meanUpdateGaussianMulti}
\end{align}
 Other distributions in the natural exponential family, e.g., the multinomial, have variances that are a function of the signal --- linear regression is the exception.

\textbf{The univariate signal case} covers the majority of applications encountered in practice. Here the signal $\lambda_t$, the response $y_t$, and the nuisance parameter $\phi$ are all scalars, and the predictor $\mathbf{x_t}$ is a vector,  so the update equations become:
\begin{align}
\mathbf{C_t} = & \mathbf{R_t} - \frac{\sigma^2_y(f_t)}{1 + \sigma^2_y(f_t) \mathbf{x_t'R_tx_t}} \mathbf{\big(R_tx_t\big)\big(x_t'R_t\big)}, \text{ and } \label{eq:varUpdatecanonicalNEFUnivariate} \\ 
\mathbf{m_t} =&  \mathbf{a_t}+ \mathbf{C_tx_t}  \frac{1}{\phi}\big(y_t-h(f_t)\big), \label{eq:meanUpdatecanonicalNEFUnivariate}
\end{align}
with $\sigma^2_y(f_t)$ being the variance of the response evaluated at the predicted signal $f_t$. Equation \ref{eq:varUpdatecanonicalNEFUnivariate} shows that the effect of the new observation is to reduce the covariance of the parameters by an amount proportional to $\mathbf{\big(R_tx_t\big)\big(x_t'R_t\big)}$, and a gain that gets smaller when there is more variance in the predicted signal, as captured by $\mathbf{x_t'R_tx_t},$ and larger when the response is expected to have a higher variance $\sigma^2_y(f_t)$.

Many common regression models fall in this category. \textbf{Univariate linear regression} with $y \sim \mathcal{N}(\lambda_t, \sigma^2)$ and   $$l(y_t|\lambda_t)=\frac{-1}{2\sigma^2}\big(y_t-\lambda_t\big)^2 =\frac{ \lambda_t y_t}{\sigma^2} - \frac{1}{2\sigma^2}(y^2_t + \lambda_t^2).$$ This is already in natural exponential form with $\phi=\sigma^2$, and with $\lambda_t$ playing the role of the natural parameter (i.e., the canonical link was used to map the mean of the response to the signal). Substituting $h(f_t)=f_t$ and $\sigma^2_y=\sigma^2=\phi$ in Equations \ref{eq:varUpdatecanonicalNEFUnivariate} and \ref{eq:meanUpdatecanonicalNEFUnivariate} yields the univariate Kalman filter.
\begin{align}
\mathbf{C_t} = & \mathbf{R_t} - \frac{\sigma^2}{1 + \sigma^2\mathbf{x_t'R_tx_t}} \mathbf{\big(R_tx_t\big)\big(x_t'R_t\big)}, \text{ and } \label{eq:varUpdatecanonicalGaussianUnivariate} \\ 
\mathbf{m_t} =&  \mathbf{a_t}+ \mathbf{C_tx_t}  \frac{1}{\sigma^2}\big(y_t-f_t\big). \label{eq:meanUpdatecanonicalGaussianUnivariate}
\end{align}

In \textbf{Poisson Regression} $y_t$ is a positive integer that follows a Poisson distribution with mean $e^{\lambda_t}$.  The likelihood is $l(y_t|\lambda_t)= y_t\lambda_t - e^{\lambda_t},$ which is again in natural exponential form with $\lambda_t$ as the natural parameter, and with $\phi=1$. The variance of a Poisson random variable is equal to its mean, so Equations \ref{eq:varUpdatecanonicalNEFUnivariate} and \ref{eq:meanUpdatecanonicalNEFUnivariate} become
\begin{align}
\mathbf{C_t} = & \mathbf{R_t} - \frac{e^{f_t}}{1 + e^{f_t} \mathbf{x_t'R_tx_t}} \mathbf{\big(R_tx_t\big)\big(x_t'R_t\big)}, \text{ and }  \label{eq:PoissonVarUpdate} \\ 
\mathbf{m_t} =&  \mathbf{a_t}+ \mathbf{C_tx_t} \big(y_t-e^{f_t}\big). \label{eq:PoissonMeanUpdate} 
\end{align}

In \textbf{Exponential Regression} $y_t$ is a non-negative real number that follows an exponential distribution with mean $1/\lambda_t$, so  $l(y_t|\lambda_t)= -y_t\lambda_t + log(\lambda_t),$ with mean $\frac{1}{\lambda_t}$ and variance $\frac{1}{\lambda_t^2}$. Note that here $\phi=-1$, so unlike other models here, the update in the mean is negatively proportional to the error, namely:
\begin{align}
\mathbf{C_t} = & \mathbf{R_t} - \frac{1}{f_t^2+\mathbf{x_t'R_tx_t}} \mathbf{\big(R_tx_t\big)\big(x_t'R_t\big)}, \text{ and }  \label{eq:ExponentialVarUpdate} \\ 
\mathbf{m_t} =&  \mathbf{a_t}- \mathbf{C_tx_t} \big(y_t-\frac{1}{f_t}\big). \label{eq:ExponentialMeanUpdate} 
\end{align}

In \textbf{Logistic Regression} the response is a Bernoulli random variable. It takes on the value 1 with probability $h(\lambda_t)=\frac{1}{1+e^{-\lambda_t}},$ and the value 0 with probability $1-h(\lambda_t)$. So $h(\lambda_t)$ is the response function, and its inverse $g(\mu)=\log{\biggr(\frac{\mu}{1-\mu}\biggr)}$ is the link function. The likelihood becomes 
\begin{align}
l(y_t|\lambda_t)=& \ y_t \log\big(h(\lambda_t)\big) + (1-y_t)\log\big(1-h(\lambda_t)\big) \nonumber \\
 =& \ y_t\log{\biggr(\frac{h(\lambda_t)}{1-h(\lambda_t)}\biggr)} +\log\big(1-h(\lambda_t)\big) \nonumber \\
 =& \ y_t \lambda_t +\log\big(1-h(\lambda_t)\big).
\end{align} This is again in the natural exponential family with $\phi=1$, and variance $h(\lambda_t)\big(1-h(\lambda_t)\big)$. The last equation above implies that $g()$ is indeed the canonical link. So Equations \ref{eq:varUpdatecanonicalNEFUnivariate} and \ref{eq:meanUpdatecanonicalNEFUnivariate} become
\begin{align}
\mathbf{C_t} = & \mathbf{R_t} - \frac{h(f_t)\big(1-h(f_t)\big)}{1+h(f_t)\big(1-h(f_t)\big)\mathbf{x_t'R_tx_t}} \mathbf{\big(R_tx_t\big)\big(x_t'R_t\big)}, \label{eq:LogisticVarUpdate} \\ 
\mathbf{m_t} =&  \mathbf{a_t} + \mathbf{C_tx_t} \big(y_t- h(f_t)\big). \label{eq:LogisticMeanUpdate} 
\end{align}

\section{Sketch Of Derivation} \label{sec:derivation}
We start from Equation \ref{eq:bayes}, and view the likelihood as a function of the model parameters, namely $ l(\mathbf{y_t}|\mathbf{\lambda_t})=l(\mathbf{y_t}|\mathbf{X'_t\theta_t})=l(\mathbf{y_t}|\mathbf{\theta_t})$. We then approximate $l(\mathbf{y_t}|\mathbf{\theta_t})$ about $\mathbf{\theta_t=a_t}$ via the second-order Taylor expansion:
\begin{align}
l(\mathbf{y_t}|\mathbf{\theta_t}) \approx & l(\mathbf{y_t}|\mathbf{a_t}) + \frac{\partial  l(\mathbf{y_t}|\mathbf{\theta_t=a_t})' }{\partial \mathbf{\theta_t}}\big(\mathbf{\theta_t - a_t}\big) \nonumber \\
&+ \frac{1}{2}\big(\mathbf{\theta_t - a_t}\big)' \biggr(\frac{\partial^2  l(\mathbf{y_t}|\mathbf{\theta_t=a_t}) }{\partial \mathbf{\theta_t}^2} \biggr)\big(\mathbf{\theta_t - a_t}\big). \label{eq:TaylorExp}
\end{align}
Because the signal is given by $\mathbf{\lambda_t=X_t'\theta_t},$ we have that
\begin{align}
 \frac{\partial  l(\mathbf{y_t}|\mathbf{\theta_t=a_t})}{\partial \mathbf{\theta_t}} = & \   \mathbf{X_t}\frac{\partial  l(\mathbf{y_t}|\mathbf{\lambda_t=f_t}) }{\partial \mathbf{\lambda_t}}, \text{ and} \nonumber \\
 \frac{\partial^2  l(\mathbf{y_t}|\mathbf{\theta_t=a_t}) }{\partial \mathbf{\theta_t}^2} = & \  \mathbf{X_t}  \frac{\partial^2  l(\mathbf{y_t}|\mathbf{\lambda_t=f_t})}{\partial \mathbf{\lambda_t}^2} \mathbf{X'_t}. \label{eq:derivation1}
\end{align}

We also make the second approximation that $(\mathbf{\theta_{t}}|D_{t-1}) \sim \mathcal{N}\big(\mathbf{m_{t-1}, C_{t-1}}\big)$. This approximation is what is needed to make the mathematics below work out, but could be justified in that the Gaussian distribution is the continuous distribution that has maximum entropy given a mean and covariance matrix, and these are the only known statistics of $(\mathbf{\theta_{t}}|D_{t-1})$.

Using the two approximations in Equation \ref{eq:bayes} results in $\log{\big(p\big(\mathbf{\theta_{t}}|D_t\big)\big)} = \log{\big(p\big(\mathbf{\theta_{t}}|D_{t-1}\big)\big)} +  l(\mathbf{y_t}|\mathbf{\theta_t})$ being proportional to
\begin{align}
  -\frac{1}{2}\big(\mathbf{\theta_t - a_t}\big)' \biggr(\mathbf{R_t^{-1}} - \frac{\partial^2  l(\mathbf{y_t}|\mathbf{\theta_t=a_t}) }{\partial \mathbf{\theta_t}^2}\biggr)  \big(\mathbf{\theta_t - a_t}\big) \nonumber \\
  + \frac{\partial  l(\mathbf{y_t}|\mathbf{\theta_t=a_t})' }{\partial \mathbf{\theta_t}}\big(\mathbf{\theta_t - a_t}\big) \nonumber \\
\propto    -\frac{1}{2}\biggr(\mathbf{\theta_t - m_t}\biggr)' \biggr(\mathbf{R_t^{-1}} - \frac{\partial^2  l(\mathbf{y_t}|\mathbf{\theta_t=a_t}) }{\partial \mathbf{\theta_t}^2}\biggr) \biggr(\mathbf{\theta_t - m_t}\biggr) \label{eq:derivation2b} \\
=    -\frac{1}{2}\biggr(\mathbf{\theta_t - m_t}\biggr)' \mathbf{C_t^{-1}} \biggr(\mathbf{\theta_t - m_t}\biggr) \label{eq:derivation2}
\end{align}
where
 \begin{align}
 \mathbf{C_t}=& \ \biggr(\mathbf{R_t^{-1}} - \frac{\partial^2  l(\mathbf{y_t}|\mathbf{\theta_t}) }{\partial \mathbf{\theta_t}^2}\biggr)^{-1}, \text{ and}  \nonumber \\
  \mathbf{m_t }=& \ \mathbf{ a_t + C_t} \frac{\partial  l(\mathbf{y_t}|\mathbf{\theta_t=a_t}) }{\partial \mathbf{\theta_t}}. \label{eq:derivation3}
  \end{align} Equation \ref{eq:derivation2b} follows from completing squares, e.g., see Appendix \ref{sec:completingSquares}, and the proportional sign indicates that terms independent of $\mathbf{\theta_t}$ were dropped. The result shows that under our approximations, $(\mathbf{\theta_{t}}|D_t\big) \sim \mathcal{N}\big(\mathbf{m_{t}, C_{t}}\big).$ To finish the argument, we substitute the expressions in Equation \ref{eq:derivation1} into Equation \ref{eq:derivation3}, and apply the Woodbury matrix inversion formula to the expression for $\mathbf{C_t}$ in Equation \ref{eq:derivation3} to  finally get the update Equations \ref{eq:varUpdate2} and \ref{eq:meanUpdate}.

\section{Contextual Multi-Armed Bandits} \label{contextualBandits}
The models we discuss here can be and have been applied to a wide range of situations to model, analyze and forecast univariate and multivariate time series. E.g., see Chapter 14 in \cite{durbin2012time} or \cite{harrison1999bayesian} for a range of examples. Here we apply the models discussed to the contextual multi-armed bandits scenario, where so far only univariate time series modeled through a linear or logistic regression have been considered. In the latter case, the only treatment known to us approximates the covariance matrix as diagonal. The models we have discussed enable explore/exploit algorithms for contextual multi-armed bandit scenarios where the reward depends on a multivariate response vector distributed according to the exponential family, and where the true parameters of the different arms are dynamic. We hope this broadens the situations where contextual multi-armed bandit approaches can be helpful. 

The standard setup involves a player interacting with a slot machine with $A \in \mathbb{Z}$ arms over multiple rounds. Every time an arm is played a reward gets generated. Different plays of the same arm generate different rewards, i.e., the reward is a random variable. Different arms have different and unknown reward distributions, which are a function of an observed context. At every time step, the player must use the observed context for each arm and all the history of the game to decide which arm to pull and then collect the reward. We seek algorithms that the player can use to decide what arm to play at every round in order to maximize the sum of the rewards received. These algorithms build statistical models to predict the reward for each arm based on the context, and decide how to balance exploring arms about which little is known with the exploitation of arms that have been explored enough to be predictable. The exploration/exploitation trade-off requires having a handle on the uncertainty of the predicted reward, so the models used need to predict at least the mean and variance of the reward for each arm. Real applications such as personalized news recommendations or digital advertising often have tight temporal and computational constraints per round, so the methods to update the statistical models with every outcome need to be online. 

Popular and useful model choices describe the univariate reward for each arm as a linear function of the context plus Gaussian noise (i.e., through a linear regression, e.g., see \cite{li2010contextual}), or through a logistic regression (\cite{chapelle2011empirical}). In the latter case, the algorithm that updates the model based on new observations uses a diagonal approximation of the parameter covariance matrix. In all these references, model parameters are assumed static (although their estimates change with every observation). In the non-contextual multi-armed bandit problem, recent efforts have tried to generalize the distributions for the rewards to the univariate exponential family \cite{korda2013thompson}, and as far as we know this is the first treatment for the contextual case.

We consider the following scenario. The parameters of all arms are compiled in the single parameter vector $\mathbf{\theta_t}$ that, unlike other settings, is allowed to change over time according to Equation \ref{eq:parDynamics}. Some entries in $\mathbf{\theta_t}$ correspond to parameters for a single arm, and others are parameters shared across multiple or all arms. We describe the model parameters via $(\mathbf{\theta_{t}}|D_{t-1}\big) \sim \mathcal{N}\big(\mathbf{a_{t}, R_{t}}\big),$ where $D_{t-1}$ is the history of contexts and responses seen up to and including round $t-1$. At the start of round $t$, we observe the context matrix $\mathbf{X_t}(a) \in \mathbb{R}^{c \times k}$ for each arm $a$, and combine this information with our knowledge of  $(\mathbf{\theta_{t}}|D_{t-1}\big)$ to decide which arm to play. Denote the arm played by $a(t)$, and its corresponding context matrix simply by $\mathbf{X_t},$ to make it consistent with the notation in the rest of this paper. Playing arm $a(t)$ results in a response $\mathbf{y_t}$ with a distribution in the (possibly multivariate) exponential family that depends on the context $\mathbf{X_t}.$ The relation between the response and the context is given by the dynamic GLM in Section \ref{sec:dynGLM}, so the mean of $\mathbf{y_t}$ is a function of the signal $\mathbf{\lambda_t}=\mathbf{X'_t\theta_t}.$ The response is used to update our estimates of the model parameters $(\mathbf{\theta_{t+1}}|D_{t}\big) \sim \mathcal{N}\big(\mathbf{a_{t+1}, R_{t+1}}\big)$, according to the algorithm described in Section \ref{sec:dynGLM}, to be used in round $t+1$. 

\begin{figure}
	\centerline{\includegraphics[width=7cm,height=7cm]{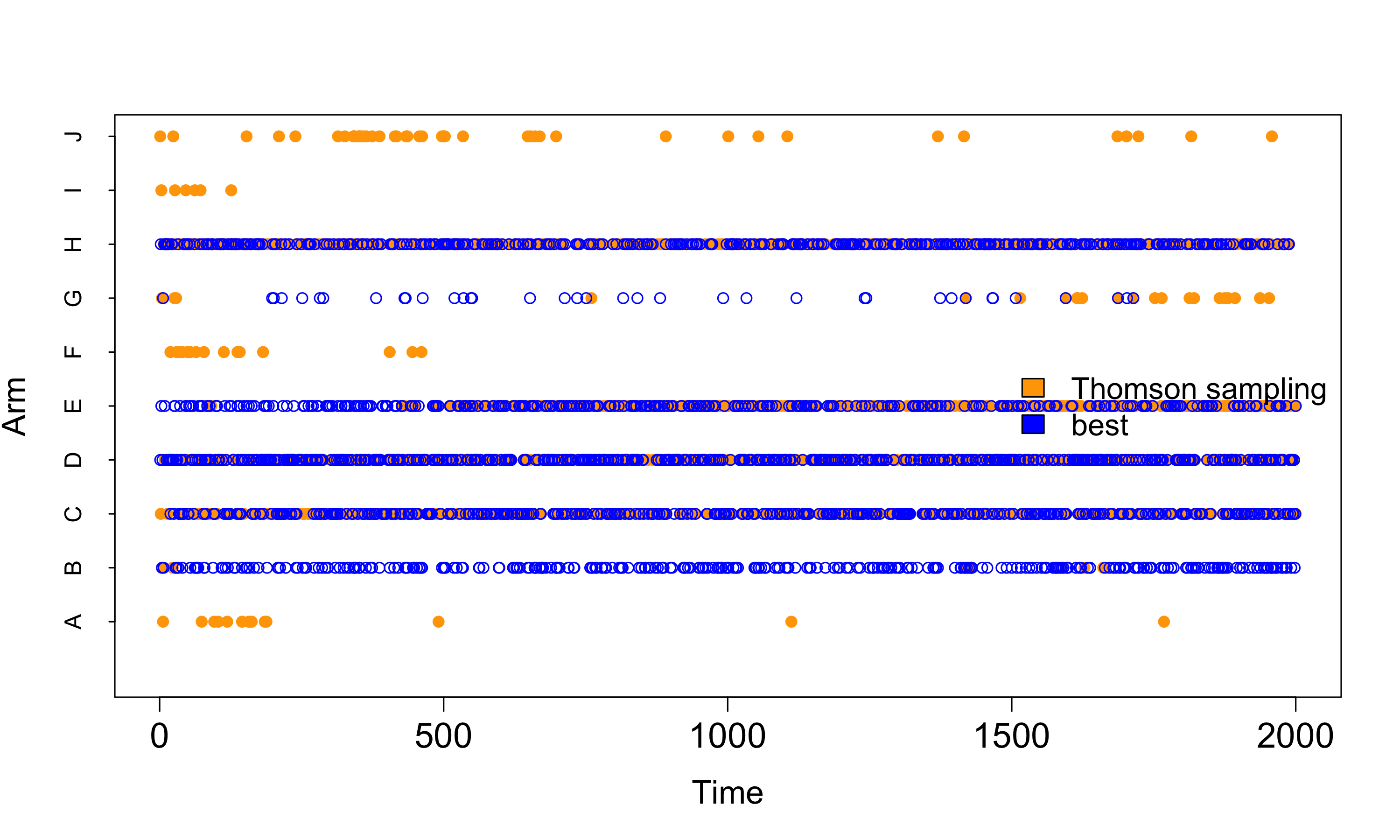}
			  \includegraphics[width=7cm,height=7cm]{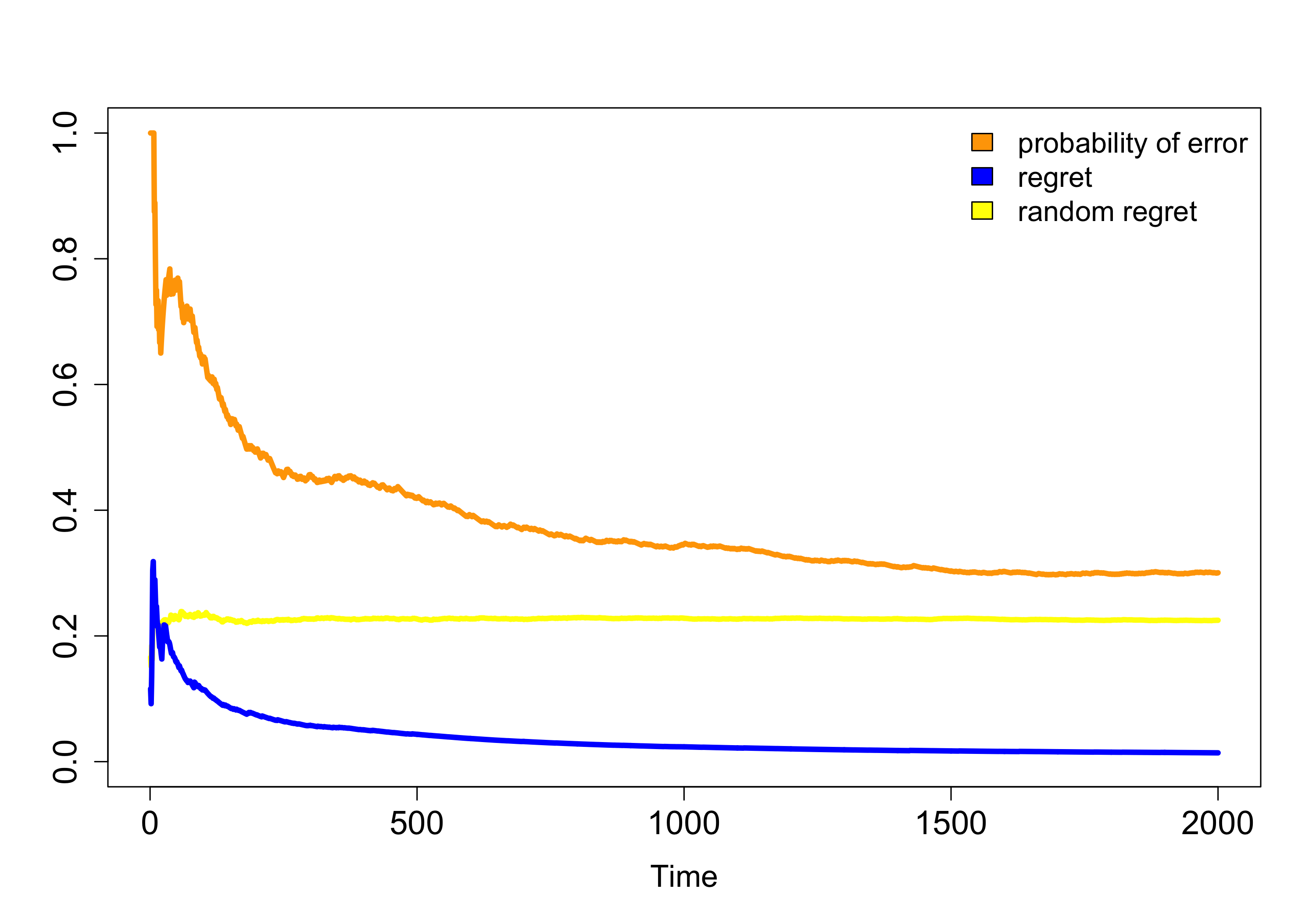}}
	\caption{(Left) Result of one simulation with 2000 rounds and 10 arms labeled A through J. The left plot shows the optimal arm in blue and the arm played in orange. (Right) The fraction of rounds where the optimal arm was not played (orange), the cumulative regret rate (blue) and the cumulative random regret rate (yellow).}
\label{fig:OneRealization}
\end{figure}

We assume the reward $r(t) = f(\mathbf{y_t})$ received in round $t$ is a known deterministic function of the response, e.g., a linear combination of the entries in $\mathbf{y_t}$. If we knew the actual model parameters, the optimal strategy to maximize the rewards collected throughout the game would be to play the arm $a^*(t)$ with the highest average reward, i.e.,  $a^*(t) =  \argmax_a E[f(\mathbf{y_t})|\mathbf{\lambda_t}=\mathbf{X'_t}(a)\mathbf{\theta_t}].$ We define the regret $\Delta(t) = E[f(\mathbf{y_t})|\mathbf{X'_t}(a^*(t))\mathbf{\theta_t}] - E[f(\mathbf{y_t})|\mathbf{\lambda_t}=\mathbf{X'_t}\mathbf{\theta_t}],$ i.e., the difference between the means of the rewards of the optimal arm and the arm played given the context and the model parameters.

Unlike the more standard contextual setup, ours allow for the explicit modeling of parameter dynamics. It also broadens the choice of probability distribution to use for the response or reward to more naturally match the model choice to the nature and dimensionality of the reward data. E.g., we can use a Poisson regression when the reward is a positive integer, or have a response with multiple entries each with a different distribution, use all response entries to update our parameter estimates, and then define the reward to be a single entry in the response.

\subsubsection{Thompson Sampling}
A contextual multi-armed bandit algorithm uses the knowledge of the parameters at each round and the context to decide which arm to play. The widely used upper confidence bound (UCB) approach constructs an upper bound on the reward for each arm using the mean and covariance of the parameter estimates at every round, and selects $a(t)$ as the arm with the highest upper bound, e.g., see \cite{li2010contextual}. Another approach that has gained recent popularity (\cite{chapelle2011empirical}, \cite{agrawal2012thompson}) is the so-called Thompson sampling introduced in \cite{thompson1933likelihood}, where arm $a$ is selected at round $t$ with a probability that it is optimal given the current distribution $(\mathbf{\theta_{t}}|D_{t-1}\big) \sim \mathcal{N}\big(\mathbf{a_{t}, R_{t}}\big)$ for the model parameters. It is only recently that asymptotic bounds for its performance have been developed both for the contextual  (\cite{agrawal2012thompson}, for the linear regression case only) and the non-contextual (\cite{agrawal2012further}) case. The studies mentioned have found Thompson sampling to perform at pair or better relative to other approaches, and to be more robust than UCB when there is a delay in observing the rewards. 

Thompson sampling is also very easy to implement. In one variant we sample a parameter value $\mathbf{\tilde{\theta}_{t}}$ from the distribution $\mathcal{N}\big(\mathbf{a_{t}, R_{t}}\big),$ and let  $a(t) =  \argmax_a E[f(\mathbf{y_t})|\mathbf{\lambda_t}=\mathbf{X'_t}(a)\mathbf{\tilde{\theta}_t}].$ In another variant, rather than sampling the model parameters once for all arms from $\mathcal{N}\big(\mathbf{a_{t}, R_{t}}\big),$ we generate independent samples from the same distribution, the sample for arm $a$ denoted by $\mathbf{\tilde{\theta}_{t}}(a)$, and then let  $a(t) =  \argmax_a E[f(\mathbf{y_t})|\mathbf{\lambda_t}=\mathbf{X'_t}(a)\mathbf{\tilde{\theta}_t}(a)].$ The latter approach is found in \cite{agrawal2012thompson} for the linear regression case to have a total regret that asymptotically scales with the number of model parameters rather than with its square as in the first variant, so we use the second variant in our simulations.

\subsection{Simulations} \label{sec:simulations}

Our goal here is to demonstrate how our online regression models work in the contextual bandits case when the observations are multivariate and not Gaussian, and when the model parameters are allowed to be dynamic. The goal is not to compare different contextual bandit algorithms, so we only focus on Thompson sampling. The model we simulate is inspired by the problem of optimizing the Netflix sign-up experience. Each arm corresponds to a variant of the sign-up pages that a visitor experiences --- a combination of text displayed, supporting images, language chosen, etc. The context corresponds to the visitor's type of device and/or browser, the day of week, time of day, country where the request originated, etc. Some of these predictors are continuous, such as the time of day, and others are categorical, such as the day of the week. The goal is maximizing signups by choosing the sign-up variant the is most likely to lead to a conversion given the context. We also observe other related outcomes associated to each visitor, such as the time spent on the sign-up experience, and whether they provide their email before signing up. We assume that these other observations are also related to the model parameters (though possibly with different context vectors), and use them to improve our parameter estimates. So our response is multivariate, even if the reward is based on a single entry of the response vector. Lastly, we want to let the model parameters drift over time, because we know that different aspects of the Netflix product are relevant over time, e.g., different videos in our streaming catalog will be the most compelling in a month than today. 

Denote the response by $\mathbf{y_t}=[y_1 \ y_2 \ y_3]',$ and the reward by $r(t) = y_1.$ We model $y_1$ through a logistic regression, with a probability of taking the value 1 of $\pi_1 = \frac{1}{e^{-\lambda_1} + 1}$ and variance $\pi_1(1-\pi_1)$. The two other response entries do not affect the reward, but we use them to improve our estimates of the model parameters. We model $y_2$ as a linear regression with mean $\lambda_2$ and variance $ \sigma^2_{y_2}=1$, and $y_3$ through another logistic regression, with mean $\pi_3 = \frac{1}{e^{-\lambda_3} + 1}$ and variance $\pi_3(1-\pi_3)$. The signal is $\mathbf{\lambda_t} = [\lambda_1 \ \lambda_2 \ \lambda_3]'=\mathbf{X'_t}\mathbf{\theta_t}.$ We assume that the entries of the response are independent of each other conditioned on the signal, so the nuisance parameter matrix $\mathbf{\Phi_t}$ is diagonal and time-independent, with the vector $[ 1 \   \sigma^2_{y_2} \ 1]'$ as its diagonal, and the covariance matrix of the response $\mathbf{\Sigma_{y_t}}$ is diagonal with the vector $[ \pi_1(1-\pi_1) \ \sigma^2_{y_2}  \ \pi_3(1-\pi_3)]'$ as its diagonal.

The context matrix $\mathbf{X_t}(a) \in \mathbb{R}^{k \times 3}$ for arm $a$ has one row for each model parameter entry and one column per response entry. Some rows correspond to parameters shared by all arms, and others to parameters corresponding to a single arm. To construct $\mathbf{X_t}(a)$ we simulate continuous and categorical predictors that we sample at every round. We let $\mathbf{X_c} \in \mathbb{R}^{k_1 \times 3}$ play the role of the continuous predictors, and sample each column from a zero-mean Gaussian with covariance $\mathbf{\Sigma_c}$. The diagonal entries in $\mathbf{\Sigma_c}$ are sampled independently from an exponential distribution with rate of 1, and the off-diagonal entries all have a correlation of $-0.1$. We let the categorical predictor $\mathbf{x_d} \in \mathbb{R}^{k_2}$ be a sample from a uniform categorical distribution with $k_2$ entries, i.e., all entries in $\mathbf{x_d}$ are zero except for one that is set to 1. We also let $\mathbf{i}(a)$ be an indicator vector that specifies that arm $a$ is being evaluated. It has $A$ entries that are all zero except for its $a$-th entry which is set to 1. Letting $\mathbf{1}_m$ be a column vector with $m$ entries, all set to 1, we define the context matrix for arm $a$ as
\begin{align}
\mathbf{X_t}(a) = & \
\begin{bmatrix} 
\mathbf{1}_3' \otimes \mathbf{i}(a) \\
 \mathbf{X_c} \\
 \mathbf{1}_3' \otimes \mathbf{x_d}\\ 
 \mathbf{i}(a) \otimes  \mathbf{X_c} \\
  \mathbf{i}(a) \otimes \big(  \mathbf{1}_3' \otimes \mathbf{x_d}\big)
\end{bmatrix}.
\end{align}
Here $\otimes$ denotes the Kronecker product between two vectors or matrices. The first $A$ rows of $\mathbf{X_t}(a)$ simply specify what arm is being evaluated, the next $k_1$ rows correspond to the continuous predictors, followed by $k_2$ rows for the categorical predictors. The next $k_1 \times A$ rows $\mathbf{i}(a) \otimes  \mathbf{X_c}$ are the interaction terms between the continuous predictors and the arm (only rows corresponding to the arm $a$ are non-zero), and the last $k_2 \times A$ rows are the interaction terms between the categorical predictor and the arm chosen (all these rows are zero except one that is set to 1). The number of rows and model parameters is then $k=A+(k_1+k_2)(A+1)$. We let $k_1=5$ and $k_2=3$. Note that without the interaction terms, the optimal arm would be independent of the context.

\begin{figure}
	\centerline{\includegraphics[width=7cm,height=7cm,keepaspectratio]{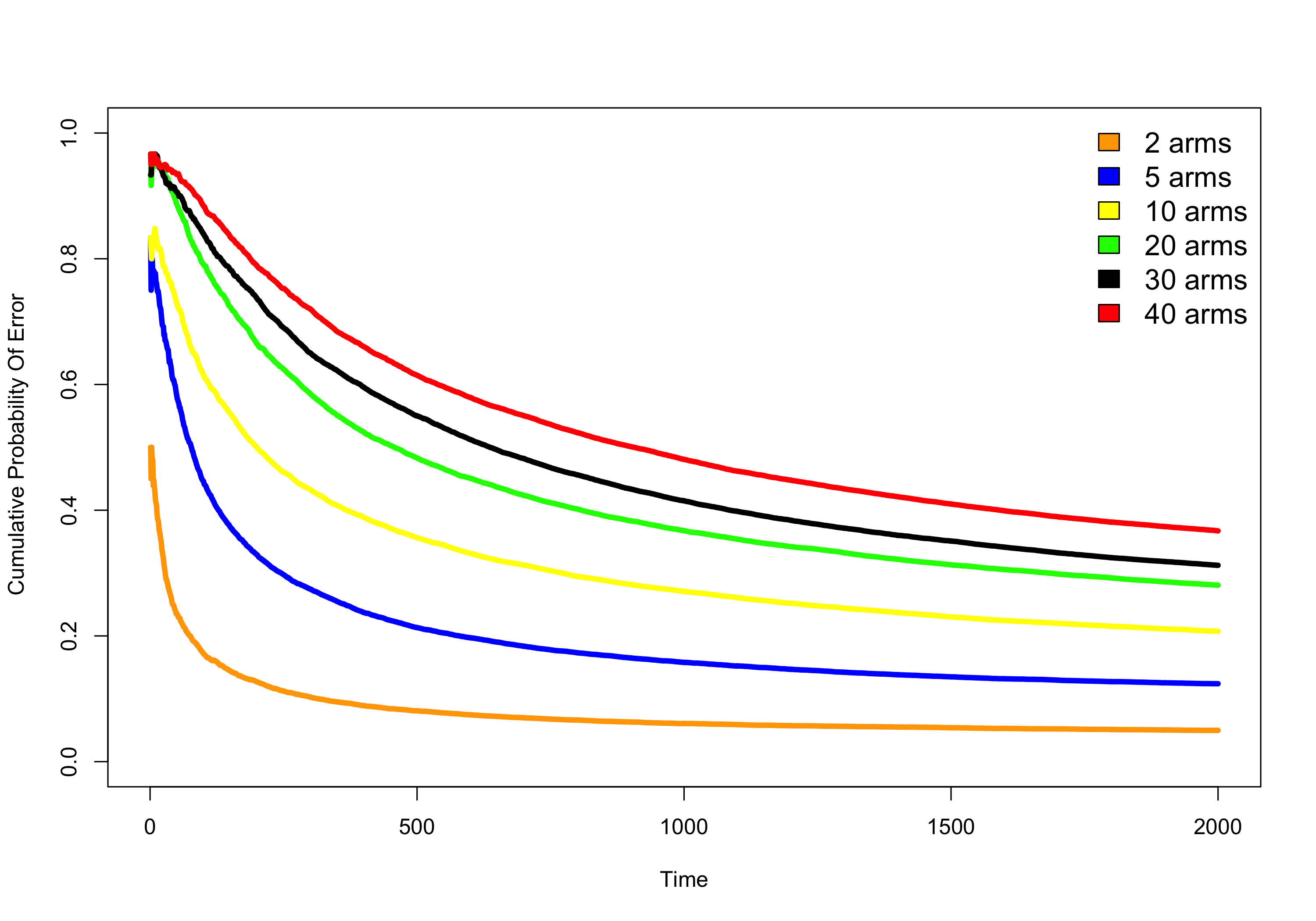}
			  \includegraphics[width=7cm,height=7cm,keepaspectratio]{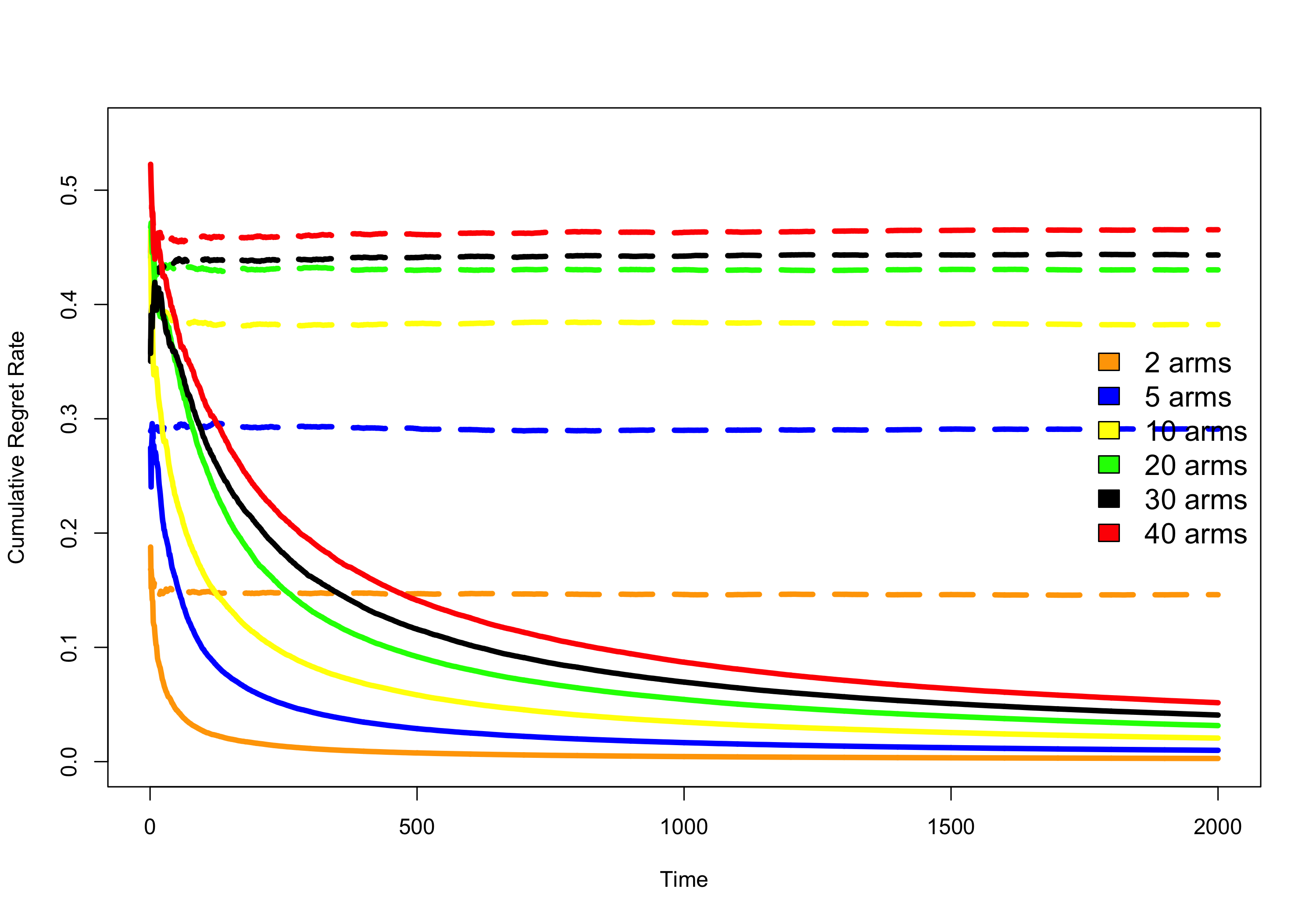}}
	\caption{(Left) Fraction of rounds when the optimal arm was not played. (Right) Cumulative regret rates (solid lines) and cumulative random regret rates (dashed lines) for scenarios with a different number of arms.}
\label{fig:AvgRegrets}
\end{figure}


We set the model parameter dynamics to $\mathbf{\theta_{t}} = \mathbf{\theta_{t-1}} + \mathbf{\omega_t},$ where $\mathbf{\omega_t} \sim \mathcal{N}\big(\mathbf{0, W_{t}}\big);$ $\mathbf{W_t}$ has diagonal entries that are independent exponential random variables with rate $c_1=10^5$, and a correlation coefficient of 0.2 for its off-diagonal entries. We sample a different matrix $\mathbf{W_t}$ at every round. We assume the first visitor arrives at $t=1$, and start the game by sampling $\mathbf{\theta_{0}}$ from a zero-mean Gaussian with diagonal covariance matrix. The diagonal entries are independent samples from an exponential distribution with rate equal to 1. We initialize the mean and covariance estimates of $\mathbf{\theta_{0}}$ as $\mathbf{m_{0}=0}$ and $\mathbf{C_{0}=I}$, where $\mathbf{I}$ is the identity matrix.

At round $t$, starting from the mean $\mathbf{m_{t-1}}$ and covariance $\mathbf{C_{t-1}}$ estimates of the parameters, we compute the mean $\mathbf{a_t}$ and covariance $\mathbf{R_t}$ of the parameters. We then sample one value of the model parameters for each arm from the resulting prior distribution, and construct the context matrices $\mathbf{X_t}(a)$ for each arm. We use the context matrices and the parameter samples to choose $a(t)$ (which defines $\mathbf{X_t}$) based on Thompson sampling, we play $a(t)$ and observe the response to obtain the round's reward, and update the parameter estimates to obtain $\mathbf{m_t}$ and covariance $\mathbf{C_t}$ and start the next round.

\begin{figure}
	\centerline{\includegraphics[width=7cm,height=7cm,keepaspectratio]{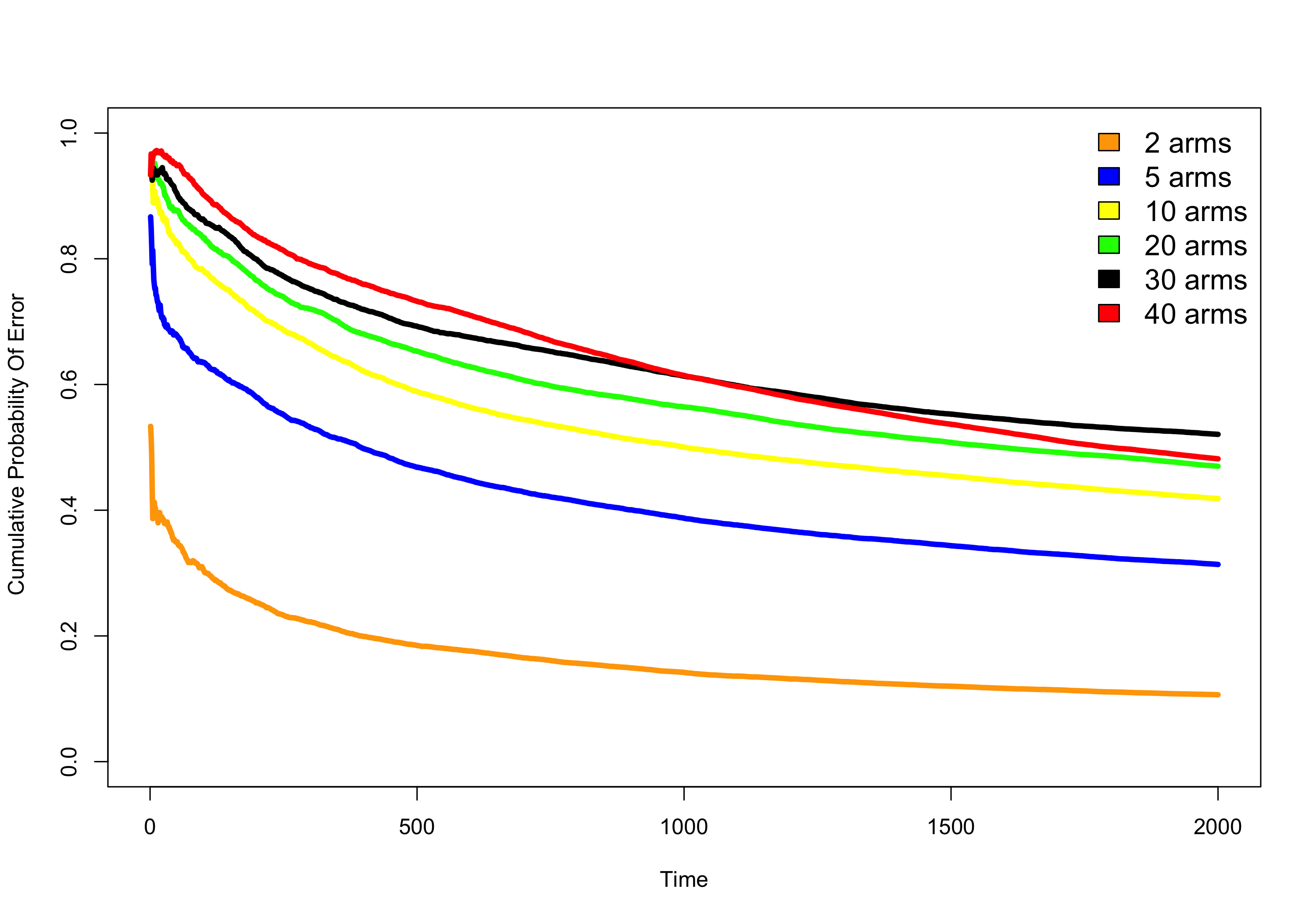}
			  \includegraphics[width=7cm,height=7cm,keepaspectratio]{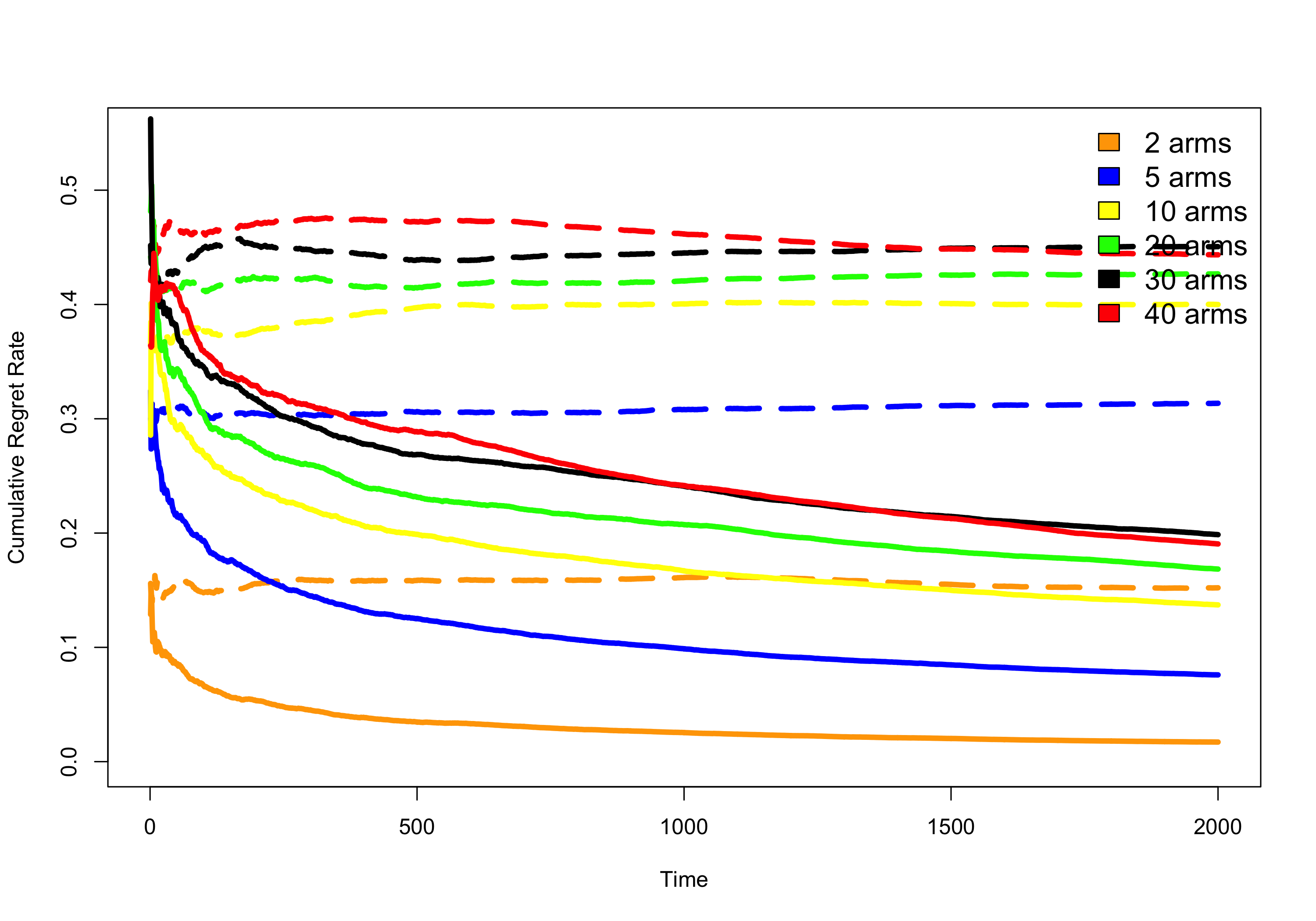}}
	\caption{This plots are equivalent to those in Figure \ref{fig:AvgRegrets}, but using $c_1=1$ rather than $c_1=10^5$  to increase the diffusion rate of the model parameters.}
\label{fig:AvgRegrets3}
\end{figure}

Figure \ref{fig:OneRealization} shows the result of one simulation with 2000 rounds and 10 arms labeled A through J. The left plot shows the optimal arm (that with the highest predicted reward $\pi_1$ based on the actual parameters $\mathbf{\theta_{t}}$) in blue and the arm played, selected via Thompson sampling and the parameter estimates, in orange. It is evident from the spread of the orange dots across arms that, as expected, there was more exploration at the start of the game. The spread of the blue dots shows that the interaction terms between the context and the arm result in different arms being optimal in different rounds. The right plot shows the fraction of rounds when the optimal arm was not chosen through the first $t$ rounds for all values of $t$ in the simulation. It drops under 0.4 from close to 1.0 at the start. The blue line on the same figure shows the cumulative average regret rate per round, which is the sum of regrets per round divided by the number of rounds. The regret per round is simply $\pi_1(a^*(t)) - \pi_1(a(t))$, both evaluated using the actual parameters $\mathbf{\theta_{t}}.$ The yellow line shows the cumulative random regret that would have resulted from choosing any arm with equal probability, independently of the model parameter estimates or the context.

We then repeated the full simulation 30 times and averaged the resulting timeseries across runs, for different scenarios with a different number of arms. Figure \ref{fig:AvgRegrets} shows the results. As expected, the probability of error, the regret and the random regret all increase with a larger number of arms. But the increased regret rate is quite mild, and continues to drop as more rounds are played. The benefit of the contextual bandit algorithm relative to uniformly at random choosing an arm is the difference between the random regret rate and the regret rate, and it increases nicely as the number of arms increases.

We expect our approach to fall apart when the parameters drift so quickly over time that the information in the observations is not enough to keep the covariance of the model parameters from growing. We explored this by increasing the parameter diffusion rate by changing $c_1$ from $10^5$ to $1.$ The results are shown in Figure \ref{fig:AvgRegrets3}: Although all metrics worsen, the regret rate still decreases nicely over time despite the large parameter fluctuations over time.

\section{Discussion} \label{conclusion}
We described a framework to easily obtain online algorithms that approximately estimate the mean and covariance matrix of the model parameters for a wide range of multivariate regression models where the model parameters change over time. Although our derivation is novel, these algorithms have been well known within a subset of the time-series community for at least a decade, but to the best of the author's knowledge, are not well known within the broader machine learning and statistical community, where we think these tools can be helpful. We also propose using the algorithms in the contextual multi-armed bandit problem, where the approach here allows for dynamic parameters and a wider range of reward distributions.

The methods we discuss here correspond to the so-called filtering problem in the Kalman filter and related literature. There are other related algorithms that solve the so-called smoothing problem, i.e., that estimate the parameters at any point in the past using all the observations. The latter have been useful for time-series analysis, but seem less obviously useful in machine learning applications (though they are well known for the standard Kalman filter, e.g., see \cite{minka1999hidden} or \cite{murphy1999filtering}), and so are not covered here. Also, in situations where the parameter dynamics are non-linear, or where higher moments of the parameter estimates are desired, there are good alternative simulation-based approaches, e.g., that rely on ideas from importance sampling and particle filters, that may be better choices than the methods described here. The best overviews of the full suite of methods that we know of are \cite{durbin2012time}, \cite{klein2003state} and \cite{harrison1999bayesian}.

\section{Acknowledgments} 
We thank Devesh Parekh and Dave Hubbard for the initial discussions that triggered this research, Stephen Boyd and George C. Verghese for the suggestion to relate this to Kalman filters, and to Justin Basilico, Roelof van Zwol, and Vijay Bharadwaj for useful feedback on this paper.

\appendix
\section{Mean And Covariance Of $(\mathbf{\theta_t}|D_{t-1})$} \label{sec:meanVarPredictionApp}
Assuming that $\big(\mathbf{\theta_{t-1}}|D_{t-1}\big) \sim \big(\mathbf{m_{t-1}}, \mathbf{C_{t-1}}\big),$ and using Equation \ref{eq:parDynamics} for the parameter dynamics, we have that
\begin{align}
E\big[\mathbf{\theta_t}|D_{t-1}\big] =& E\big[ \mathbf{G_t\theta_{t-1}} + \mathbf{B_t u_{t-1}}+\mathbf{\omega_t}|D_{t-1}\big] \nonumber \\
=&   \mathbf{G_t m_{t-1}} + \mathbf{B_t u_{t-1}} = \mathbf{a_t},
\end{align}
resulting in Equation \ref{eq:meanPred}. Here we used the assumption that $E\big[\mathbf{\omega_t}|D_{t-1}\big] = \mathbf{0}$. 

The covariance matrix $\mathbf{R_t}$ of $(\mathbf{\theta_t}|D_{t-1})$ in Equation \ref{eq:varPred} is found as follows:
\begin{align}
\mathbf{R_t} =& E\big[(\mathbf{\theta_t-a_t})(\mathbf{\theta_t-a_t})'|D_{t-1}\big] \nonumber \\
=& E\big[\big(\mathbf{G_t}(\mathbf{\theta_{t-1}}  -\mathbf{m_{t-1}})+ \mathbf{\omega_t}\big)\big(\mathbf{G_t}(\mathbf{\theta_{t-1}}  -\mathbf{m_{t-1}})+ \mathbf{\omega_t}\big)'    |D_{t-1}\big] \nonumber \\
=& \mathbf{G_tC_{t-1}G_t'}+\mathbf{W_t}.
\end{align}
Here, we used the assumption that the noise vector $\mathbf{\omega_t}$ is uncorrelated with the parameter $\mathbf{\theta_{t-1}}.$

The mean and covariance $\mathbf{\Omega_t} = E[(\mathbf{\lambda_t}- \mathbf{f_t})(\mathbf{\lambda_t}- \mathbf{f_t})'|D_{t-1},\mathbf{X_t}] $ of the signal are derived as follows:
\begin{align}
E[\mathbf{\lambda_t}|D_{t-1},\mathbf{X_t}] =& \mathbf{X_t'}E[\mathbf{\theta_t}|D_{t-1}] =  \mathbf{X_t'a_t}=\mathbf{f_t}. \nonumber \\
 \mathbf{\Omega_t}=& E\big[\mathbf{X_t'}(\mathbf{\theta_t - a_t})(\mathbf{\theta_t - a_t})' \mathbf{X_t} | D_{t-1},\mathbf{X_t} \big] \nonumber \\
=& \mathbf{X_t'R_tX_t} .
\end{align}
So $\big(\mathbf{\lambda_{t}}|D_{t-1}, \mathbf{X_t}\big) \sim \big(\mathbf{f_t, \Omega_t}\big)$.

Lastly, the covariance $E\big[(\mathbf{\lambda_t-f_t})(\mathbf{\theta_t-a_t})'|D_{t-1},  \mathbf{X_t}\big]$ between the signal and the parameters at time $t$ given $D_{t-1}$ and the predictors $\mathbf{X_t}$ is given by
\begin{align}
 \mathbf{X_t'} E\big[(\mathbf{\theta_t-a_t})(\mathbf{\theta_t-a_t})'|D_{t-1}\big] = \mathbf{X_t'R_t}.
\end{align}

\section{The Exponential Family} \label{sec:meanVarNEF}
Let $\mathbf{y}$ be a random vector with $d$ entries distributed according to the exponential form 
\begin{align}
l \big( \mathbf{y} | \mathbf{\eta}, \mathbf{\Phi}\big) = & \mathbf{\eta' \Phi^{-1} T(y)} - b\big(\mathbf{\eta, \Phi} \big)+c\big(\mathbf{y, \Phi} \big), \label{eq:ef}
\end{align}
where $\mathbf{T(y)}$ is a sufficient statistic for $\mathbf{y}$, $\mathbf{\eta}$ is the natural parameter vector, and $\mathbf{\Phi}$ is a symmetric $d$-by-$d$ matrix and a nuisance parameter. Note that Equation \ref{eq:nef} is more restrictive, because it implicitly assumes that  $\mathbf{T(y)}= \mathbf{y}$.
\subsubsection{Example: Gaussian Distribution}
For example, if $\mathbf{y} \sim \mathcal{N}\big(\mathbf{\mu, \Sigma}\big),$ with known covariance matrix $\mathbf{ \Sigma}$ but unknown mean, we have that $l \big( \mathbf{y} | \mathbf{\eta}, \mathbf{\Phi}\big)$ is
\begin{align}
  - \frac{1}{2}\biggr((\mathbf{y}-\mathbf{\mu})'\mathbf{\Sigma^{-1}}(\mathbf{y}-\mathbf{\mu}) \biggr)  - \frac{1}{2}\log(|\Sigma|) -\frac{k}{2}\log(2\pi) \nonumber \\
= \underbrace{ \mathbf{\mu}'\mathbf{\Sigma^{-1}}\mathbf{y} }_{\mathbf{\eta' \Phi^{-1} T(y)}} - \underbrace{\frac{1}{2} \mathbf{\mu}'\mathbf{\Sigma^{-1}} \mathbf{\mu}}_{b\big(\mathbf{\eta, \Phi} \big)}  \underbrace{- \frac{1}{2}\mathbf{y'}\mathbf{\Sigma^{-1}}\mathbf{y} - \frac{1}{2}\log(|\Sigma|) -\frac{k}{2}\log(2\pi)}_{c\big(\mathbf{y, \Phi} \big)}, \label{eq:GaussianNEF}
\end{align}
so $\mathbf{y}$ in in the exponential family with  $\mathbf{\Phi}=\mathbf{\Sigma}$,  $\mathbf{\mu}$ is the natural parameter $ \mathbf{\eta}$, and the sufficient statistic is $\mathbf{T(y)}=\mathbf{y}$. When the covariance matrix is unknown, the sufficient statistic expands to include  $\mathbf{yy'},$ and $\mathbf{\eta}$ is a function of both $\mathbf{\mu}$ and $\mathbf{\Sigma}$.

If the covariance matrix is not known, then we can instead define the sufficient statistic to be $\mathbf{T(y)'}=[\mathbf{y'} \ vec(\mathbf{yy'})'],$ where $ vec(\mathbf{A})$ for any matrix $\mathbf{A}$ is a column vector resulting from stacking all the columns in $\mathbf{A}$, and re-arrange Equation \ref{eq:GaussianNEF} to have the natural exponential form in Equation \ref{eq:ef}, now with natural parameter $\mathbf{\eta}$ being both a function of $\mathbf{\mu}$ and $\mathbf{\Sigma},$ with $l \big( \mathbf{y} | \mathbf{\eta}, \mathbf{\Phi}\big) $ is proportional to:
\begin{align}
\mathbf{\mu}'\mathbf{\Sigma^{-1}}\mathbf{y} - \frac{1}{2}\mathbf{y'}\mathbf{\Sigma^{-1}}\mathbf{y} - \frac{1}{2} \mathbf{\mu}'\mathbf{\Sigma^{-1}} \mathbf{\mu}  - \frac{1}{2}\log(|\Sigma|) =\nonumber \\
 \underbrace{[\mathbf{\mu}'\mathbf{\Sigma^{-1}} \ -\frac{1}{2}vec(\Sigma^{-1})']}_{\mathbf{\eta}'}\underbrace{[\mathbf{y'} \ vec(\mathbf{yy'})]'}_{\mathbf{T(y)}} - \underbrace{ \big(\frac{1}{2} \mathbf{\mu}'\mathbf{\Sigma^{-1}} \mathbf{\mu}  + \frac{1}{2}\log(|\Sigma|)\big)}_{b(\mathbf{\eta})} ,\label{eq:GaussianNEF2}
\end{align}
so in this case $\mathbf{\Phi}=\mathbf{I},$ and the natural parameter becomes a function of $\mathbf{\Sigma}$ as well as of the mean  $\mathbf{\mu}$.

\subsubsection{Moment Generating Function}
The function $M_{\mathbf{y}}(\mathbf{t}) =  E[\exp{\big(\mathbf{t'\Phi^{-1}T(y)}\big)}]$ can be shown to equal $\exp{\big(b(\mathbf{\eta+t, \Phi})-b(\mathbf{\eta, \Phi}) \big)}$.
\begin{align}
 \int \exp{(\mathbf{t'\Phi^{-1} y})} \exp{\biggr( \mathbf{\eta' \Phi^{-1} T(y)} - b\big(\mathbf{\eta, \Phi} \big)+c\big(\mathbf{y, \Phi} \big)\biggr)} d\mathbf{y} = \nonumber \\
 \int  \exp{\biggr( \mathbf{(t+\eta)' \Phi^{-1} T(y)} - b\big(\mathbf{t+\eta, \Phi} \big)+c\big(\mathbf{y, \Phi} \big)\biggr)} d\mathbf{y}  \nonumber \nonumber \\ \times \biggr(\exp{\big(b(\mathbf{\eta+t, \Phi})-b(\mathbf{\eta, \Phi}) \big)} \biggr) =  \exp{\big(b(\mathbf{\eta+t, \Phi})-b(\mathbf{\eta, \Phi}) \big)}. \label{eq:mgf}
\end{align}
The last equality follows because the integrand in the first term of the second equation is also a probability distribution in the natural exponential family with parameter $\mathbf{t+\eta}$, so the integral equals 1.
Taking the first derivative of $M_{\mathbf{y}}(\mathbf{t})$ and evaluating it at $\mathbf{t}=0$ yields
\begin{align}
\frac{dM_{\mathbf{y}}(\mathbf{t=0})}{d\mathbf{t}} &=   \mathbf{\Phi^{-1}}E[\mathbf{T(y)}] = \frac{\partial b(\mathbf{\eta, \Phi})}{\partial \mathbf{\eta}}. \label{eq:mgfMean}
\end{align}
Similarly, the second derivative of $M_{\mathbf{y}}(\mathbf{t})$ at $\mathbf{t}=0$ yields
\begin{align}
\mathbf{\Phi^{-1}}E[\mathbf{T(y)T(y)'}]  \mathbf{\Phi^{-1}} =  \frac{\partial b(\mathbf{\eta, \Phi})}{\partial \mathbf{\eta}} \biggr( \frac{\partial b(\mathbf{\eta, \Phi})}{\partial \mathbf{\eta}}\biggr)' +  \frac{\partial^2 b(\mathbf{\eta, \Phi})}{\partial \mathbf{\eta^2}}, \text{ so}  \nonumber  \\
 \mathbf{\Phi} \frac{\partial^2 b(\mathbf{\eta, \Phi})}{\partial \mathbf{\eta^2}}  \mathbf{\Phi} =E\biggr[\biggr(\mathbf{T(y)-E[\mathbf{T(y)}]}\biggr)\biggr(\mathbf{T(y)-E[\mathbf{T(y)}]}\biggr)'\biggr]. \label{eq:mgfVar}
\end{align}
Setting $\mathbf{T(y)=y}$ above yields Equations \ref{eq:nefMean} and \ref{eq:nefVar}.

\section{Completing Squares For Quadratic Matrix Functions} \label{sec:completingSquares}
The derivation of Equation \ref{eq:derivation2} required turning the expression $\mathbf{a'Ca+b'a},$ where $\mathbf{a}=\big(\mathbf{\theta_t - a_t}\big),$ $\mathbf{C}= \mathbf{R_t^{-1}} - \frac{\partial^2  l(\mathbf{y_t}|\mathbf{\theta_t}) }{\partial \mathbf{\theta_t}^2}$ (which is symmetric and positive definite) and $ \mathbf{b}= \frac{\partial  l(\mathbf{y_t}|\mathbf{\theta_t}) }{\partial \mathbf{\theta_t}}$ into the so-called perfect square expression $\mathbf{(a-h)'C(a-h)},$ for $\mathbf{h=m_t}$.
We have $\mathbf{(a-h)'C(a-h)} =  \mathbf{a'Ca} -2\mathbf{h'Ca} + \mathbf{h'Ch},$  so we need $-2\mathbf{h'Ca} = \mathbf{b'a}.$ This implies that $\mathbf{h'}=\mathbf{-b'C^{-1}/2}.$

\end{document}